\documentclass[Afour,sageh,times]{sagej}

\usepackage[colorlinks,bookmarksopen,bookmarksnumbered,citecolor=red,urlcolor=red]{hyperref}

\usepackage{amsmath,amsfonts}
\usepackage{array}
\usepackage[caption=false,font=normalsize,labelfont=sf,textfont=sf]{subfig}
\usepackage{textcomp}
\usepackage{stfloats}
\usepackage{url}
\usepackage{verbatim}
\usepackage{graphicx}

\usepackage{amsmath}
\usepackage{amsthm}
\usepackage{algorithm}
\usepackage[noend]{algpseudocode}
\usepackage{amssymb}  
\usepackage[dvipsnames, table]{xcolor}
\usepackage{graphicx}
\usepackage{booktabs}
\usepackage{hyperref}
\usepackage{mathtools}
\usepackage{multirow}
\usepackage{siunitx}

\DeclareMathOperator*{\minimize}{minimize}				
\DeclareMathOperator*{\maximize}{maximize}				
\DeclareMathOperator*{\st}{subject\,to}					

\newcommand{\Xs}[0]{\ensuremath{\mathcal{X}}}						
\newcommand{\Us}[0]{\ensuremath{\mathcal{U}}}						
\newcommand{\Ss}[0]{\ensuremath{\mathcal{S}}}						
\newcommand{\Bs}[0]{\ensuremath{\mathcal{B}}}						

\newcommand{\T}[0]{\ensuremath{\top}}							
\newcommand{\Rv}[1]{\ensuremath{\mathbb{R}^{#1}}}				
\newcommand{\Expect}{{\rm I\kern-.3em E}}				
\newcommand{\norm}[1]{\left\lVert#1\right\rVert}        

\newtheorem{theorem}{Theorem}
\newtheorem{lemma}{Lemma}

\newtheorem{assumption}{Assumption}
\newtheorem{definition}{Definition}

\makeatletter
\newenvironment{proof}[1][\proofname]{\par
\pushQED{\qed}%
\normalfont \topsep6\p@\@plus6\p@\relax
\trivlist
\item\relax
{\itshape
#1\@addpunct{.}}\hspace\labelsep\ignorespaces
}{%
\popQED\endtrivlist\@endpefalse
}
\makeatother

\newenvironment{eqs*}{\begin{equation}\aligned}{\endaligned\end{equation}}
\newenvironment{eqs}[1]{\equation\label{#1}\aligned}{\endaligned\endequation}

\graphicspath{{figures/}}

\providecommand{\review}[1]{#1}   

\algnewcommand\And{\textbf{and} }
\algnewcommand\Or{\textbf{or} }

\def\volumeyear{2026}

\begin{document}


\runninghead{Lunardi, Fontanari, Saveriano and Del Prete}

\title{Beyond Terminal Constraints:\\ Reliable Robot Control with Learned Safe Sets}

\author{Gianni Lunardi\affilnum{1,2}, Elias Fontanari\affilnum{2}, Matteo Saveriano\affilnum{2}, and Andrea Del Prete\affilnum{2}}

\affiliation{%
\affilnum{1}Faculty of Engineering, Free University of Bolzano, Bolzano, Italy\\
\affilnum{2}Department of Industrial Engineering, University of Trento, Trento, Italy}

\corrauth{Gianni Lunardi and Elias Fontanari.}

\email{gianni.lunardi@unibz.it, elias.fontanari@unitn.it}




\begin{abstract}
Ensuring safety in advanced robot control methods such as Model Predictive Control (MPC) and Reinforcement Learning remains an open challenge. A common approach relies on control-invariant sets, yet their exact computation for nonlinear systems is generally intractable, and numerical approximations compromise theoretical guarantees. In this work, we show that safety does not need to rely exclusively on control invariance. We introduce novel classes of safe sets, each broader than classical control-invariant sets, and \review{analyze} corresponding MPC schemes that guarantee recursive feasibility by constraining different nodes of the prediction horizon rather than only the terminal state. Furthermore, when the safe set is an inner approximation of a suitable control-invariant set, our methods can ensure safety through a safe-abort strategy, even in the absence of recursive feasibility. We validate our approaches in extensive simulations on a 6-joint robot manipulator, showing fewer constraint violations than state-of-the-art methods while maintaining comparable tracking performance and computational efficiency. Hardware experiments further demonstrate the practicality of our schemes for real-world robotic applications.

\end{abstract}

\keywords{Model Predictive Control, Safety in Robotics, Control Invariance}
\maketitle

\section{Introduction}

\label{sec:introduction}

Safety is a fundamental concern in robotics, playing a crucial role in enabling reliable, robust, and trustworthy operations. Safety concerns are multifaceted, including obstacle avoidance, prevention of falls in legged robots, and avoidance of joint and actuator limits, which could otherwise lead to hardware damage or system failure. 

Among the many challenges in ensuring robotic safety, two areas stand out due to their importance and relevance: Human-Robot Interaction (HRI) and safe learning control. In HRI~\cite{Zanchettin2016, Villani2018}, robots must operate in proximity to humans, necessitating strict guarantees to avoid collisions and ensure human well-being. This requires advanced perception, planning, and control systems capable of responding rapidly to unpredictable human behavior while adhering to safety standards. 

\begin{figure}[tb]
    \centering
    \includegraphics[width=\columnwidth]{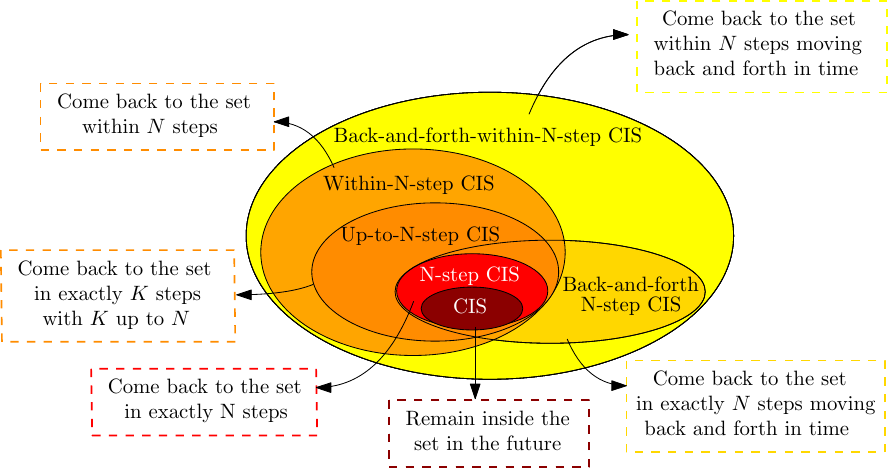}
    \caption{\review{Hierarchy of safe set classes proposed in this work. The innermost set represents the well-known class of Control Invariant Sets (CISs). We introduce broader classes, namely up-to-$N$-step CISs, within-$N$-step CISs, back-and-forth-$N$-step CISs, and back-and-forth-within-$N$-step CISs.}}
    \label{fig:safe_sets}
\end{figure}

In parallel, the advent of learning-based control methods~\cite{Hwangbo2019, Rudin2022} has revolutionized the field of robotics, enabling robots to adapt to complex and uncertain environments. However, the safety of learning-based control methods remains a critical concern, as they often lack formal guarantees and can exhibit unsafe behaviors during training or deployment. Ensuring safety of learning-based control methods is essential for their widespread adoption in real-world applications.

These challenges motivate the need for advanced control methods that can provide formal safety guarantees while enabling robots to operate in complex and dynamic environments. In the literature, there are three main tools to certify the safety of a controller: Control-Invariant Sets (CISs)~\cite{Blanchini1999, Grune2017}, Control Barrier Functions (CBFs)~\cite{Ames2014, Ames2017, Wu2019}, \review{and predictive shielding methods based on} Back-Up Policies (BUPs)~\cite{Bastani2021}. These tools are typically applied to different control methods, respectively Model Predictive Control (MPC), Quadratic-Programming (QP) based control, and Reinforcement Learning (RL). 

Despite the great amount of work devoted to safety in the last decade, ensuring safety for general high-dimensional nonlinear systems is still an open problem.
A key challenge seems to be the computation of so-called CISs (a.k.a. safe sets).

\subsection{Contributions}
\review{In this paper, we introduce new classes of safe sets that are broader than CISs. Indeed, as shown in Fig.~\ref{fig:safe_sets}, these novel classes contain the class of CISs, and rely on the concept of multi-step safety, which allows the state to leave the set for a limited number of time steps before entering it again. }
These sets generalize the concept of CIS, meaning that they must satisfy weaker requirements than CISs.
We hypothesize that these new sets might therefore be easier to compute and certify with numerical approximation methods. This is confirmed by empirical evaluations on the computed safe sets,  even though we do not thoroughly validate this hypothesis in this paper. 
\review{Instead, we show that three existing MPC formulations (\emph{Receding-Constraint MPC}, \emph{Hard Terminal-Constraint MPC}, and \emph{Parallel-Constraint MPC}) can ensure RF or safety assuming the safe set belongs to a specific class. Unlike classic approaches that only use the safe set as a terminal constraint, these MPC schemes apply the constraint at different nodes along the prediction horizon. 
With respect to our previous work~\cite{Lunardi2024, Fontanari2025}, we present the following extensions spanning theory, computational efficiency, and validation:}
\review{
\begin{itemize}
\item We present \emph{four classes of safe sets} that contain the class of CISs, of which only one was discussed in our previous work.
\item We correct a mistake in the theory of \emph{Receding-Constraint MPC}~\cite{Lunardi2024}, clarifying the property of the safe set that it needs to ensure RF.
\item We prove that \emph{Parallel-Constraint MPC} can ensure RF under weaker conditions than \emph{Receding-Constraint MPC}, defining the associated novel class of safe sets (back-and-forth-within-N-step CISs).
\item We prove that the classic \emph{Hard Terminal-Constraint MPC} can ensure safety (without RF) even if the terminal constraint set is not a CIS, but it belongs to the larger class of back-and-forth-$N$-step CISs.
\item While in our previous work the safe-abort procedure was too slow for a real-time implementation, this limitation is mitigated through a suitable choice of the safe set and the optimization algorithm. 
\item While our previous work had only considered joint-space constraints, we consider here more realistic scenarios featuring Cartesian obstacles and self-collision constraints. Moreover, our open-source software implementation\footnote{\url{https://github.com/idra-lab/safe-mpc}.} integrates a URDF-based dynamics library (\textsc{Adam}\footnote{\url{https://github.com/gbionics/adam}.}), which makes our software easily adaptable to any robotic system.
\item We present a broad simulation-based statistical analysis comparing several MPC formulations (novel and classic) in different working conditions (initial conditions, horizon lengths, sizes of the safe set, modeling errors).
\item We report on the first implementation of \emph{Receding-Constraint MPC}~\cite{Lunardi2024} on hardware, which validates the practical applicability of the approach on a full 6-DOF robot manipulator.
\end{itemize}
}
\review{
The rest of the paper is structured as follows. In Section~\ref{sec:related_work}, we review prior research addressing safety challenges in robotics. Section~\ref{sec:problem} introduces the nominal MPC formulation and related concepts. Section~\ref{sec:safe_mpc} presents the analyzed MPC formulations together with the respective safe-set definitions and the proofs of RF. Sections~\ref{sec:task_abortion} and~\ref{sec:safe_set_computation} provide details on the safe-abort procedure and the computation of the safe sets. Section~\ref{sec:software} discusses the gap between the developed theory and the practical implementation. In Section~\ref{sec:simulations}, we validate our approaches through simulations on a robotic manipulator. Then, we show the applicability of Receding-Constraint MPC on the real robot in Section~\ref{sec:experiments}. Finally, Section~\ref{sec:conclusions} draws the conclusions.
}

\section{Related Work}
\label{sec:related_work}

\subsection{\review{MPC and Control Invariant Sets}}
\label{ssec:mpc_ci}
MPC has emerged as a promising approach to address safety challenges. Indeed, given its optimization nature, MPC enables the incorporation of safety constraints within the control framework. This allows robots to operate safely in complex environments while adapting to changing conditions and uncertainties. In practice, MPC for constrained nonlinear systems still strives to persistently satisfy constraints, a problem referred to as \emph{Recursive Feasibility} (RF). 
Indeed, while constraint satisfaction is ensured for infinite-horizon MPC, this is in general computationally intractable~\cite[Chapter 2]{Diehl2017book}. In practice, RF is achieved with a long enough horizon~\cite[Chapter 8]{Grune2017}, but in general we cannot know how long it should be, and even if we knew it, it may result in unacceptable computation time. 
For finite horizon MPC, RF can be guaranteed if a CIS is used to constrain the terminal state of the MPC horizon~\cite{Mayne2000}. Still, CISs are generally unknown for nonlinear systems, which include most robotic systems. Trivial CISs often exist (e.g., the equilibria of the system), and are known. However, these CISs are small and result in poor performance, e.g., by making the system move slowly to reach zero velocity by the end of the horizon.    
Different numerical techniques exist to approximate larger CISs~\cite{Djeridane2006, Coquelin2007, Jiang2016, Rubies2016, Hsu2021, Dawson2023, Zhou2020, LaRocca2023, Leung2020}, and they typically rely on function approximators, such as deep Neural Networks (NNs). However, all safety guarantees are lost if the computed set is not exactly control invariant, which cannot be guaranteed in general by these numerical methods. 
\review{Moreover, even an under-approximation of the viability kernel (i.e., the maximal CIS) is itself not necessarily control invariant.}
The most common approaches for certifying approximated CISs are based either on Lipschitz constant estimates~\cite{Boffi2021} or on a \emph{learner-verifier} architecture combined with optimization methods~\cite{Dai2020}. Both approaches, however, face significant limitations: estimating Lipschitz constants can be difficult and often provides overly conservative bounds, while also suffering from the curse of dimensionality; learner-verifier methods require solving optimization problems whose computational cost grows exponentially with the number of neurons in the NN used to approximate the CIS.

\subsection{\review{Control Barrier Functions (CBF)}}
CBFs have emerged as an alternative to CISs: indeed, these two frameworks are strongly connected, since the zero-sublevel set of a CBF defines a CIS.
CBFs are often used in safety filters: a desired control input is minimally modified through a QP to satisfy the CBF decrease condition~\cite{Wabersich2023}. 
In~\cite{Xu2015}, the authors prove the robustness of CBFs in presence of model uncertainties.
This type of controller has been deployed in different robotic contexts, from mobile robots to manipulators and quadrupeds~\cite{Singletary2022, Jian2023, Grandia2021}. Nonetheless, similarly to CISs, CBFs do not have a general constructive method for nonlinear systems: practitioners tried to synthesize CBFs using Sum-of-Squares Programming~\cite{Wang2023}, for which analytical certifications exist, or learning techniques~\cite{Dawson2022,Dawson2023, Robey2020}, but still it is difficult to verify the satisfaction of Nagumo's condition~\cite{Nagumo1942} on the whole state space for high-dimensional systems. 

\subsection{\review{Model Predictive Shielding (MPS)}}
A different approach is MPS~\cite{Bastani2021, Bastani2021acc, Wabersich2022}, which exploits a BUP that tries to drive the system to a known CIS. 
Contrary to MPC, MPS can achieve good performance even if this CIS is extremely small, thanks to the use of the BUP.
Before applying the control computed by the nominal policy, the safety of the next state is verified by simulating the dynamics with the BUP. If the state reaches the CIS, then this certifies the safety of the next state, and the nominal policy can be used. Otherwise, the BUP is used at that time step, and the process is repeated at the next iteration. 
Contrary to CISs and CBFs, BUPs do not require certification, since their ability to reach the CIS is checked at each control loop.  
In addition, MPS does not demand the system dynamics to be differentiable, as the BUP is known in advance and simply applied at runtime. 
However, MPS requires simulating the system dynamics for potentially long horizons, which can be too computationally demanding for real-time applications.
Moreover, since the nominal policy is unaware of safety, this can lead to frequent use of the BUP, resulting in poor performance.



\subsection{\review{Hamilton-Jacobi Reachability (HJR)}}
Safety has also been tackled with HJR analysis, which is closely related to control invariance theory and CBFs. 
HJR~\cite{Bansal2017} solves the Hamilton–Jacobi–Isaacs partial differential equation to compute finite or infinite-horizon Backward Reachable Tubes (BRTs), i.e. sets of states from which system trajectories inevitably reach a target set within the given time horizon, regardless of the control input. If the target set encodes constraint violations, its complement is a CIS in the infinite-horizon setting.
Typically, HJR approaches take into account disturbances acting on system dynamics: indeed, safety-critical scenarios are represented as zero-sum differential games~\cite{Bansal2017}, where Player A is the control input that moves the system towards the target, while Player B models the disturbance that pushes the system to violate the safety constraints.
HJR is based on Dynamic Programming theory, thus it suffers from the curse of dimensionality. Nonetheless, recent learning approaches have been able to mitigate this issue: in~\cite{Bansal2021}, the authors proposed \emph{DeepReach}, a method that exploits sinusoidal NN to solve high-dimensional reachability problems leveraging GPUs. In~\cite{So2024}, the relationship between HJR and CBF has been exploited to develop a novel class of neural CBFs, which has been tested with state spaces up to 16D. Another method to compute infinite-time BRTs is the Viability-Boundary Optimal Control (VBOC) algorithm~\cite{LaRocca2023}: VBOC finds states directly on the border of the viability kernel, thus requires less sampling than HJR and CBF methods. Nevertheless, all these methods only obtain approximate CISs and, due to their learning-based nature, do not offer any formal guarantees. Moreover, VBOC is applicable only to fully actuated systems.

\subsection{\review{Multi-Step Control Invariance}}
While all the works discussed so far revolved around classic control invariance, in~\cite{Zhao2024}, the authors proposed a novel class of safe sets: $K$-step CISs. These sets have the property that from every state inside them it is possible to reach the set in at most $K$ steps. In other words, $K$-step CISs allow the state to exit the set for at most $K-1$ steps. Using this type of sets, they developed a novel MPC scheme that guarantees RF. Assuming an MPC formulation with an $N$-step horizon, they solve $K$ MPC problems with different horizons (from $N+1$ to $N+K$) and a $K$-step CIS as terminal constraint; then they prove that at least one of the problems is guaranteed to be feasible. However, this results in long-horizon problems that can be demanding to solve in fast control loops. 

A similar approach has been proposed in~\cite{Lunardi2024}, where an $N$-step CIS constraint moves backward along the horizon at each MPC iteration. Moreover, the safe CIS is also used as a soft terminal constraint to ensure persistent RF. Those concepts have been pushed further in~\cite{Fontanari2025}, where (up to) $N$ MPC problems are solved in parallel, each of them having the safe-set constraint on a different node of the horizon. Differently from~\cite{Zhao2024}, the last two methods are more amenable to real-time implementation as they do not require increasing the horizon length.

\subsection{\review{Safety for Robotic Manipulation}}
\review{
Certifying the safety of motion planning algorithms in HRI is critical. A popular approach to safe manipulation is CHOMP~\cite{Zucker2013}, which iteratively optimizes the trajectory starting from an initial guess by using function gradient techniques, making a trade-off between smoothness and obstacle avoidance. 
Safety is mostly guaranteed through costs and distance fields rather than control invariance or reachability analysis. In \cite{Holmes2020}, a set of parameterized reachable trajectories is precomputed offline for each manipulator joint and then combined online to form conservative reachable sets in the workspace, which serve as collision-avoidance constraints within a receding-horizon optimizer. 
\cite{Michaux2025} extend this idea from kinematic planning to uncertain manipulator dynamics by combining polynomial-zonotope reachable sets with a robust passivity-based controller.
In collaborative robotics, safety has also been studied through speed and separation monitoring~\cite{Scalera2024}, where the supervisory controller guides the robot along a path-consistent stop trajectory when collision danger between human and robot is detected. Also, the size of the robot safety is adapted based on the stopping time of the manipulator.
}

\review{
Overall, these works show that manipulator safety is often handled by exploiting platform-specific structure, such as serial kinematics, swept-volume approximations, simple geometric primitives, and braking maneuvers, rather than applying generic safe-control tools directly. 
}
\section{Background}
\label{sec:problem}
This section introduces the notation used in this paper, the problem of \emph{safety} in MPC, together with the concepts of Control-Invariant Set (CIS) and Recursive Feasibility (RF).

\subsection{Notation}
Throughout this work, the following notation will be used:
\begin{itemize}
    \item $\mathbb{N}, \Rv{}$ stand for the sets of natural and real numbers, respectively;
    \item $I$ represents the identity matrix \review{(its dimension can be inferred by the context)};
    \item $x$ represents the system state, while $u$ denotes the control input;
    \item $x[:m]$ indicates the first $m$ elements of the vector $x$, while $x[m:]$ denotes all elements of $x$ starting from the $(m+1)$-th one up to the last one;
    \item $\{ x_i \}_0^N$ designates a discrete-time trajectory, namely the sequence $(x_0, \dots, x_N)$, \review{where the single subscript $i$ indexes the internal time step within the sequence};
    \item $x_{i|k}$ indicates the state predicted at step $k+i$, obtained when solving the MPC optimization problem at time $k$; \review{when the conditioning index $k$ is omitted, the remaining subscript refers to the time step $i$};
    \item $\Xs$ and $\Us(x)$ are respectively the feasible state and control sets, where the control set is in general state dependent;
    \item$\mathcal{S} \subseteq \Xs$ is the safe set used by the MPC formulation to ensure safety;
    \item the symbol $.^*$ is used for the optimal values;
    \item the symbol $.^g$ is used for the solver guess (i.e., previous MPC solution).
\end{itemize}

\subsection{Problem Statement}
Let us examine a discrete-time dynamical system subject to state constraints and (state-dependent) control constraints:
\begin{equation} \label{eq:sys_dynamics}
    x_{i+1} = f(x_i, u_i),
\qquad
    x \in \Xs, \qquad u \in \Us(x) .
\end{equation}
\review{The set $\Xs$ represents the set of admissible states. In particular, it includes both the intrinsic state bounds of the system and state constraints arising from the environment, such as collision-avoidance constraints from static obstacles.}
The objective is to design a control strategy that guarantees \emph{safety} (i.e., constraint satisfaction), while at the same time retaining performance (i.e., minimizing the cost) as much as possible.
We denote by $\mathcal{E}$ the set of the equilibrium states of the system, namely:
\begin{equation}
    \mathcal{E} = \{ x \in \Xs\ | \, \exists \, u \in \Us(x) : x = f(x,u) \}.
\end{equation}
To accomplish our goal, we make use of the concepts of $M$\emph{-step Backward-Reachable Set} and \emph{Control Invariant Set}~\cite{Blanchini1999}.
\begin{definition}
    \label{def:brs}
    Given a set $\mathcal{A} \subseteq \Xs$, the \underline{$M$-step Backward Reachable Set} (BRS) of $\mathcal{A}$ is defined as the subset of $\Xs$ starting from which it is possible to reach $\mathcal{A}$ in $M$ time steps \review{without leaving \Xs}: 
    \begin{equation}
    \begin{aligned}
    \Bs_M(\mathcal{A}) \triangleq \{ &x_0 \in \Xs \, | \, \exists \{u_i\}_{0}^{M-1}: x_{M} \in \mathcal{A},\\
    & x_i \in \Xs, u_i \in \Us(x_i), \forall \, i=0,\dots,M-1 \} .
    \end{aligned}
    \end{equation}
\end{definition}
\begin{definition}
    \label{def:cis}
    Given a set $\mathcal{A} \subseteq \Xs$, we say that $\mathcal{A}$ is a \underline{Control Invariant Set} if and only if: 
    \begin{equation}
    \begin{aligned}
    \forall x \in \mathcal{A}, \,\, \exists \, u \in \Us(x): f(x,u) \in \mathcal{A}.
    \end{aligned}
    \end{equation}
\end{definition}

As all BRSs of equilibrium states, the set $\Bs_M(\mathcal{E})$ is a CIS~\cite{Blanchini1999}.
Namely, starting inside $\Bs_M(\mathcal{E})$, it is possible to remain inside $\Bs_M(\mathcal{E})$ indefinitely because we can reach $\mathcal{E}$ and then remain in $\mathcal{E}$ forever.
If we knew $\Bs_M(\mathcal{E})$ (or any other CIS) we could use it to construct a safe controller. 
Now we discuss how the knowledge of a CIS can be exploited in an MPC formulation to achieve safety.

\subsection{MPC and Recursive Feasibility (RF)}
\label{ssec:recursive_feasibility}
Let us consider the following MPC problem:
\begin{subequations}
\begin{align}
        \minimize_{\{x_i\}_0^N,\{u_i\}_0^{N-1}}  &\quad \sum_{i = 0}^{N-1} \ell_i(x_i,u_i) + \ell_N(x_N),  \label{eq:mpc_cost} \\
        \st &\quad x_0 = x_{\mathrm{init}}, \label{eq:mpc_initial_conditions}\\
            &\quad x_{i+1} = f(x_i, u_i), \; \; \forall i = 0 \dots N-1  \label{eq:mpc_dynamics}\\
            &\quad x_i \in \Xs, u_i \in \Us(x_i), \; \; \forall i = 0 \dots N-1 \label{eq:mpc_bound_constraints}\\
            &\quad x_{N} \in \mathcal{X}_N, \label{eq:mpc_terminal_constraint}
\end{align}
\label{eq:mpc}
\end{subequations}
where $\ell_i(\cdot)$/$\ell_N(\cdot)$ is the running/terminal cost, $x_{\mathrm{init}}$ is the current state, and $\mathcal{X}_N \subseteq \Xs$ is the terminal set~\cite{Mayne2000}.

A standard approach to enforce RF in MPC consists in introducing a terminal set $\Xs_N$ that constrains the final state to lie within a CIS, for instance $\Bs_M(\mathcal{E})$. Unfortunately, we cannot assume to know $\Bs_M(\mathcal{E})$ in general, since CISs are notably difficult (and often impossible) to compute for nonlinear systems. As recalled in Section~\ref{ssec:mpc_ci}, numerical techniques are available to approximate $\Bs_M(\mathcal{E})$, yet an approximate set $\Ss \approx \Bs_M(\mathcal{E})$ is generally not control invariant.
Therefore, adopting $\Ss$ as terminal set in the MPC formulation does not guarantee RF.
A widely used strategy to handle the potential loss of feasibility is to soften the terminal constraint by introducing a slack variable, which is heavily penalized in the cost function~\cite{Kerrigan2000soft, Zeilinger2014}.
In this way, even when the terminal constraint cannot be met exactly, the optimization still provides a solution that enables control of the system, with the expectation that feasibility of the terminal constraint will be eventually recovered.
Nonetheless, this approach does not guarantee \emph{safety} or RF,
since the softening allows the state to exit the set $\Ss$, which eventually may cause constraint violations.

\section{Safe Model Predictive Control}
\label{sec:safe_mpc}

This section introduces the analyzed MPC schemes, namely the Receding-Constraint MPC, the Hard Terminal-Constraint MPC, and the Parallel-Constraint MPC. These controllers are presented together with the novel safe sets that are necessary to guarantee RF or safety. 

\subsection{Receding-Constraint MPC (RC-MPC)} 
\label{ssec:rc_mpc}
\begin{figure}[tb]
    \centering
    \includegraphics[width=\columnwidth]{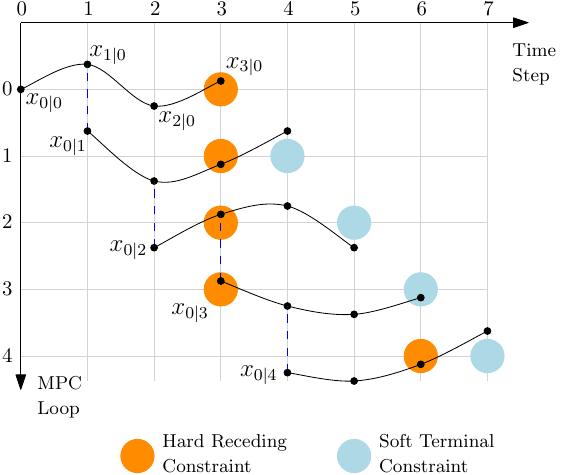}
    \caption{
        Example of RC-MPC with $N=3$ and a 3-step CIS. 
        At MPC iteration 3, the condition $x_{3|0} \in \Ss$ implies, from~\eqref{eq:n_step_control_invariance}, that the state can return to $\Ss$ in 3 steps. The soft terminal constraint $x_{3|3} \in \Ss$ is therefore feasible and can be satisfied. Consequently, the receding constraint shifts forward to time step 2.
    }  
    \label{fig:receding_constr_example}
\end{figure}

RC-MPC~\cite{Lunardi2024} is based on the concept of \emph{receding constraint}: instead of relying exclusively on the final state to ensure safety, RC-MPC exploits the fact that, as long as at least one state $x_r \in \Ss$ (with $1 \le r \le N$), we know that $x_1$ is safe because from $x_1$ we can reach $x_r$. 
RC-MPC dynamically adjusts online the time step at which the state is required to lie in $\Ss$.
For example, if during iteration $k-1$ of the MPC scheme the condition $x_{r|k-1} \in \Ss$ was satisfied, then at iteration $k$ we can infer (under the assumption of no disturbances or modeling errors) that $x_{r-1|k} \in \Ss$ is feasible, and thus we can impose it as a hard constraint.
This mechanism ensures safety for $r$ consecutive iterations, with the receding constraint shifting backward along the prediction horizon.
Once the constraint has receded all the way to step 0, however, we can no longer rely on it to ensure safety.
For this reason, RC-MPC also incorporates a soft constraint on the terminal state, encouraging it to remain inside $\Ss$. A simple example highlighting the working principle of RC-MPC is depicted in Fig.~\ref{fig:receding_constr_example}, that presents a corrected version of a similar scheme in~\cite{Lunardi2024}.

This MPC scheme is summarized in Alg.~\ref{alg:receding} and it relies on the following OCP:
\begin{equation}
    \begin{aligned}
        \minimize_{\{x_i\}_0^N,\{u_i\}_0^{N-1}, s}  &\quad \sum_{i = 0}^{N-1} \ell_i(x_i,u_i) + \ell_N(x_N) + w_s |s| \\
        \st &\quad \eqref{eq:mpc_initial_conditions}, \eqref{eq:mpc_dynamics}, \eqref{eq:mpc_bound_constraints}\\
            & \quad x_r \in \Ss \\
            &\quad x_{N} \in \Ss \oplus s\, \mathbb B, \\
    \end{aligned}
    \label{eq:rc_mpc}
\end{equation}
where the L1 penalty, \review{with weight $w_s$}, is applied to the slack variable \review{$s \ge 0$, $\oplus$ denotes the Minkowski sum, and $\mathbb B$ is the unit ball.}
After completing the MPC optimization at iteration $k-1$, we verify whether $x_{N|k-1} \in \Ss$. If this condition holds, then at cycle $k$ the receding constraint can be shifted to $x_{N-1|k}$, which guarantees safety for additional $N-1$ iterations. 
Even if not theoretically necessary, to improve the empirical performance we actually check whether $x_j \in \Ss$ for any $j>r$ (Line 6 in Alg.~\ref{alg:receding}), and if that is the case we set $r=j-1$. \review{The last optimal solution is shifted by one time step and used as initial guess for the next MPC iteration (Line 9).}

\begin{algorithm}[t]
\caption{Receding-Constraint MPC}
\small
\begin{algorithmic}[1]
\Require 
Number of time steps $T$,
Initial state $x_0$,
\mbox{Initial guess $\{x^g_i\}_0^{N}, \{u^g_i\}_0^{N-1}$},
$\Ss$ up-to-N-step CIS,
OCP~\eqref{eq:rc_mpc}
    \State $r \leftarrow N$  \Comment{Receding constraint index}
    \For{$t = 0 \rightarrow T-1$}
        \State $\{x_i^*\}_0^{N}, \{u_i^*\}_0^{N-1} \leftarrow \textsc{OCP}(r, x_t, \{x^g_i\}_0^{N}, \{u^g_i\}_0^{N-1})$
        \For{$j = r+1 \rightarrow N$} \Comment{Search last state in $\Ss$}
            \If{$x_j^* \in \Ss$}
                \State $r \leftarrow j$
            \EndIf
        \EndFor
        \State $r \gets r - 1$  \Comment{Recede constraint}
        \State $x_{t+1} \leftarrow f(x_t,u_0^*)$ \Comment{Simulate dynamics}
        \review{\State $\{x^g_i\}_0^{N}, \{u^g_i\}_0^{N-1} \leftarrow \textsc{WarmStart}(\{x_i^*\}_1^{N}, \{u_i^*\}_1^{N-1})$}
    \EndFor
\end{algorithmic}
\label{alg:receding}
\end{algorithm}



To clarify the theoretical properties of RC-MPC, we introduce two novel classes of multi-step CISs.
\begin{definition}
    Given a natural number $K\ge 1$, a set $\Ss \subseteq \Xs$ is \underline{$K$-step control invariant}\footnote{Despite using the same name, this definition is different from the one provided in~\cite{Zhao2024}.} if, starting from any state in $\Ss$, it is possible to reach $\Ss$ in exactly $K$ time steps:
    \begin{equation} \begin{aligned} \label{eq:n_step_control_invariance}
        \forall x_0 \in \Ss: &\, 
        \exists \, \{ u_i \}_{0}^{K-1}, x_K \in \Ss,\\
        & x_i \in \Xs, \,
        u_i \in \Us(x_i), \,
        \forall i=0, \dots, K-1
    \end{aligned} \end{equation}
    Using the notation introduced above for BRS, this condition can also be expressed as: $\Ss \subseteq \Bs_K(\Ss)$.
\end{definition}
This notion generalizes the classical concept of control invariance, where the special case of 1-step control invariance coincides with the standard definition.
We now introduce two classes of safe sets.
\begin{definition}
A set $\Ss \subseteq \Xs$ is \underline{up-to-$N$-step control invariant} if there exists a value $K \in [1, N]$ such that $\Ss$ is $K$-step control invariant.
    This condition can be expressed as: $$\exists \, K \in [1, N]: \quad \Ss \subseteq \Bs_K(\Ss)$$
\end{definition}
Notice that the class of up-to-$N$-step CIS contains all $K$-step CIS for all $K$ going from 1 to $N$.
\begin{definition}
    \label{def:within_n_step_cis}
    A set $\Ss \subseteq \Xs$ is \underline{within-$N$-step control invariant}\footnote{This definition is equivalent to the one used in~\cite{Lunardi2024}, where it was simply called $N$-step CIS.} if, starting from any state in $\Ss$, it is possible either to remain in $\Ss$, or to leave $\Ss$ and come back to it within $N$ time steps:
    \begin{equation} \begin{aligned} \label{eq:within_n_step_control_invariance}
        \forall x_0 \in \Ss: &\, 
        \exists \, \{ u_i \}_{0}^{k-1}, \,
        1 \le k \le N, x_k \in \Ss,\\
        & x_i \in \Xs, \,
        u_i \in \Us(x_i), \,
        \forall i=0, \dots, k-1
    \end{aligned} \end{equation}
    Using the notation introduced above for BRS, this condition can also be expressed as:
    $\Ss \subseteq \bigcup_{j=1}^N \Bs_j(\Ss).$
\end{definition}
\review{This definition is an extension of the previous one; indeed, any up-to-$N$-step CIS is also a within-$N$-step CIS. 
The difference between these set classes is that for up-to-$N$-step CISs the number of steps to return to the set is the same for all states, while for within-$N$-step CISs this number can be different for different states.}

Now we can state the conditions for the recursive feasibility of RC-MPC~\eqref{eq:rc_mpc}.

\begin{theorem} \label{th:recursive_feasibility}
Consider the RC-MPC formulation~\eqref{eq:rc_mpc}. Assuming that the set $\Ss$ is up-to-$N$-step control invariant, and the penalty on the soft terminal constraint $w_s$ is larger than the associated optimal Lagrange multiplier, then the RC-MPC described in Alg.~\ref{alg:receding} is recursively feasible.
\end{theorem}
\begin{proof}
    Suppose that the RC-MPC is feasible at the initial iteration $k=0$, which means that $x_{N|0}^* \in \Ss$.
    In this case, RF is ensured for at least $N$ subsequent iterations, during which the receding constraint can shift backwards along the horizon.
    Throughout these iterations, two possible scenarios can occur. (i) While receding $r$ the condition $x_j^* \in \Ss$ (Line 5 of Alg.~\ref{alg:receding}) is satisfied for a certain $j \in [r + 1, N]$. Then the receding constraint moves forward to time step $j$. This scenario is possible, however it is not guaranteed to occur.
    (ii) Since the set $\Ss$ is up-to-$N$-step control invariant by assumption, we know it exists $K \in [1, N]$ such that $\Ss$ is $K$-step control invariant. 
    Suppose the receding constraint reaches node $r = N-K$,  since $\Ss$ is $K$-step control invariant, the soft terminal constraint is feasible because condition~\eqref{eq:n_step_control_invariance} ensures it exists a state trajectory starting from $x_{N-K | K}^* \in \Ss$ reaching $\Ss$ in $K$ time steps.
    Assuming that the weight $w_s$ is sufficiently large~\cite{Nocedal2006}, the adopted L1 penalty function becomes exact. As a consequence, we can infer that the soft terminal constraint will be exactly satisfied.
    Once this occurs, the hard receding constraint can be moved to step $N-1$, which guarantees RF for additional $N-1$ iterations.
    By repeating the same reasoning, RF can then be maintained indefinitely.
\end{proof}

This theorem highlights how RC-MPC guarantees RF even if the set $\Ss$ is not control invariant. 
We rely indeed on a weaker condition, which is up-to-$N$-step control invariance. 
Our condition is weaker because any $1$-step CIS is also up-to-$N$-step control invariant, therefore our approach guarantees RF for a larger class of sets, which contains the class of CISs.

\subsubsection{Comparison with \cite{Lunardi2024}}
In previous work~\cite{Lunardi2024}, within-$N$-step CIS (Def.~\ref{def:within_n_step_cis}) was incorrectly used for proving the RF of RC-MPC. 
The error stems from the implicit assumption that the state satisfying the receding constraint $x_r$ remains the same throughout the MPC iterations. If this were the case, the proof in~\cite{Lunardi2024} would be correct. However, since this is not the case in general, it can happen that $x_r$ changes throughout the MPC iterations in such a way that always prevents reaching $\Ss$ from $x_r$ in $N-r$ time steps.

\subsection{Hard Terminal-Constraint MPC (HTC-MPC)}
\label{ssec:htc_mpc}
Let us consider a classic Hard Terminal-Constraint MPC (HTC-MPC) formulation with terminal constraint set $\Ss$:
\begin{equation}
    \begin{aligned}
        \minimize_{\{x_i\}_0^N,\{u_i\}_0^{N-1}}  &\quad \sum_{i = 0}^{N-1} \ell_i(x_i,u_i) + \ell_N(x_N) \\
        \st &\quad \eqref{eq:mpc_initial_conditions}, \eqref{eq:mpc_dynamics}, \eqref{eq:mpc_bound_constraints} \\
            &\quad x_{N} \in \Ss.  \\
    \end{aligned}
    \label{eq:htc_mpc}
\end{equation}
As discussed in Section~\ref{sec:problem}, we cannot guarantee RF of problem~\eqref{eq:htc_mpc} if $\Ss$ is not a CIS. 
However, when problem~\eqref{eq:htc_mpc}  is unfeasible, we can use the last computed solution while we keep trying to solve the problem at each MPC iteration. 
This idea is detailed in Alg.~\ref{alg:terminal}, and in the following we prove that it leads to safety under a weaker condition than control invariance.
\begin{algorithm}[t]
    \caption{Hard Terminal-Constraint MPC}
    \small
    \begin{algorithmic}[1]
    \Require 
    Number of time steps $T$,
    Initial state $x_0$,
    Initial guess $\{x^g_i\}_0^{N}, \{u^g_i\}_0^{N-1}$,
    $\Ss$ back-and-forth-N-step CIS,
    OCP~\eqref{eq:htc_mpc}
        \For{$t = 0 \rightarrow T-1$}
            \State $\{x_i^*\}_0^{N}, \{u_i^*\}_0^{N-1}, feas \leftarrow \textsc{OCP}(x_t, \{x^g_i\}_0^{N}, \{u^g_i\}_0^{N-1})$
            \If{$feas$ = False} \Comment{If OCP is unfeasible}
		\State \Comment{Copy last feasible solution}
                \State $\{x^*_i\}_0^{N}, \{u^*_i\}_0^{N-1} \leftarrow \{x_i^g\}_0^{N}, \{u_i^g\}_0^{N-1}$ 
            \EndIf
            \State $x_{t+1} \leftarrow f(x_t,u_0^*)$ \Comment{Simulate system}
            \review{\State $\{x^g_i\}_0^{N}, \{u^g_i\}_0^{N-1} \leftarrow \textsc{WarmStart}(\{x_i^*\}_1^{N}, \{u_i^*\}_1^{N-1})$}
        \EndFor
    \end{algorithmic}
    \label{alg:terminal}
    \end{algorithm}    

\begin{definition}
    A set $\Ss$ is \underline{back-and-forth-$N$-step control} \underline{invariant} if it satisfies the following condition:
    \begin{equation} \label{eq:condition_htcmpc}
    \exists \, k \in [0, N-1]: \quad \Bs_k(\Ss) \subseteq \Bs_N(\Ss).
    \end{equation}
\end{definition}
The reason for the name \emph{back-and-forth} is that this condition requires that from every state $x_k \in \Ss$ it must be possible to move backward in time of $k$ time steps (inside $\Bs_k(\Ss)$) and then forward in time of $N$ time steps to reach $\Ss$.

\begin{theorem} \label{th:within_recursive_feasibility}
Assuming $\Ss$ is back-and-forth-$N$-step control invariant, the HTC-MPC described in Alg.~\ref{alg:terminal} is safe. 
\end{theorem}
\begin{proof}
Assume the HTC-MPC is feasible at the first MPC iteration, which implies that $x_{N|0} \in \Ss$.
Without loss of generality, suppose that at the MPC iteration 1 the OCP gets unfeasible, thus the terminal constraint cannot be satisfied. In this scenario, we use the last solution to control the system and we keep trying to solve the OCP~\eqref{eq:htc_mpc}. After $k$ steps of unfeasibility, we will have $x_{N-k|k} \in \Ss$, since $x_{N-k|k} =x_{N|0}$. 
This implies that $x_{0|k} \in \mathcal{B}_{N-k}(\Ss)$.
Since $\Ss$ is a back-and-forth-$N$-step CIS by assumption, we know that in one of those $k \in [1, N]$ cycles we must have: $\mathcal{B}_{N-k}(\Ss) \subseteq \Bs_N(\Ss)$. 
This implies that $x_{0|k} \in \Bs_N(\Ss)$, and therefore the terminal constraint $x_{N|k} \in \Ss$ must be feasible.
When this happens, the solution ensures safety for another $N-1$ iterations.
At this point the same reasoning can be applied to ensure safety indefinitely.
\end{proof}
This theorem proves the safety of HTC-MPC with a back-and-forth-$N$-Step CIS, when used as described in Alg.~\ref{alg:terminal}.
However, this safety comes at the cost of potentially controlling the system in open-loop for up to N time steps.
This could lead to poor performance in practice, and is therefore a drawback with respect to the other two MPC formulations discussed in this work, which can instead ensure RF.

\subsection{Parallel-Constraint MPC (PC-MPC)}
\label{sec:pc_mpc}
Starting from the idea of RC-MPC (Section~\ref{ssec:rc_mpc}), instead of using a single state along the horizon to ensure safety, Parallel-Constraint MPC (PC-MPC)~\cite{Fontanari2025} tries to reach the safe set at different time steps exploiting parallel computation. 
As long as at least one state $x_p \in \Ss$ (with $1 \le p \le N$), we can ensure constraint satisfaction because $x_1 \in \Bs_{p-1}(\Ss)$.
This is because from $x_1$ we can reach $x_p$ in $p-1$ steps. 
%
\review{If parallel computation is available}, we can solve $N$ problems in parallel, each constraining the state inside $\Ss$ at a different time step $p \in [1, N]$:
\begin{equation}
    \begin{aligned}
        \minimize_{\{x_i\}_0^N,\{u_i\}_0^{N-1}}   &\quad \sum_{i = 0}^{N-1} \ell_i(x_i,u_i) + \ell_N(x_N)\\
        \st &\quad \eqref{eq:mpc_initial_conditions}, \eqref{eq:mpc_dynamics}, \eqref{eq:mpc_bound_constraints} \\
            & \quad x_p \in \Ss
    \end{aligned}
    \label{eq:parallel_constr_mpc}
\end{equation}
Then, we can use the solution of one of the problems that have been solved.
Since our main concern is safety, we decide to use the solution satisfying the safe set constraint at the furthest time.
A pseudo-code is available in Alg.~\ref{alg:parallel_mpc}.
Now we focus on the theoretical properties of PC-MPC, starting with a novel sufficient condition on the safe set $\Ss$ to ensure RF.

\begin{algorithm}[t]
\caption{Parallel-Constraint MPC}
\small
\begin{algorithmic}[1]
\Require 
Number of time steps $T$,
Initial state $x_0$,
Initial guess $\{x^g_i\}_0^{N}, \{u^g_i\}_0^{N-1}$,
$\Ss$ back-and-forth-within-N-step CIS,
OCP~\eqref{eq:parallel_constr_mpc}
    \State $N_{\mathrm{safe}} \leftarrow N$  \Comment{Safe horizon}
    \For{$t = 0 \rightarrow T-1$}
        \For{$p = 1 \rightarrow N$}
            \State $\{x^p_i\}_0^N, \{u^p_i\}_0^{N-1} \leftarrow \textsc{OCP}(p, x_t, \{x^g_i\}_0^{N}, \{u^g_i\}_0^{N-1})$ 
            \State $r_p \leftarrow -1$
            \For{$j = N_{\mathrm{safe}} \rightarrow N$} \Comment{Search last state in $\Ss$}
                \If{$x^p_j \in \Ss$}
                    \State $r_p \leftarrow j$
                \EndIf
            \EndFor
        \EndFor
        \State $p \leftarrow \Call{ArgMax}{\{r_p\}_1^N}$ \Comment{Find best solution}
        \State $N_{\mathrm{safe}} \leftarrow p - 1$\label{line:N_safe_parallel}
        \State $x_{t+1} \leftarrow f(x_t,u_0^p)$ \Comment{Simulate dynamics}
        \review{\State $\{x^g_i\}_0^{N}, \{u^g_i\}_0^{N-1} \leftarrow \textsc{WarmStart}(\{x_i^p\}_1^{N}, \{u_i^p\}_1^{N-1})$}
    \EndFor
\end{algorithmic}
\label{alg:parallel_mpc}
\end{algorithm}
\begin{definition}
    A set $\Ss$ is \underline{back-and-forth-within-$N$-step control invariant} if it satisfies the following condition:
    \begin{equation} \label{eq:condition_rf_pcmpc}
    \exists \, r \in [0, N-1]: \quad \Bs_r(\Ss) \subseteq \bigcup_{j=r+1}^N \Bs_j(\Ss).
    \end{equation}
\end{definition}
This condition requires that from every state $x_r \in \Ss$ it must be possible to move backward in time of $r$ time steps (inside $\Bs_r(\Ss)$) and then forward in time of $j$ time steps (with $j \in [r+1, N]$) to reach a state $x_j \in \Ss$.

\begin{theorem}
    \label{th:bf_recursive_feasibility}
    Assuming $\Ss$ is back-and-forth-within-$N$-step control invariant, the PC-MPC formulation described in Alg.~\ref{alg:parallel_mpc} with OCP~\eqref{eq:parallel_constr_mpc} is recursively feasible.
\end{theorem}
\begin{proof}
    The safe horizon length $N_{\mathrm{safe}}$ is initialized to $N$ and it decreases by at most 1 per control loop.
    To prove this theorem, we show that every time $N_{\mathrm{safe}}$ takes the value $r$, it is increased to a value $j\ge r$.
    This proves recursive feasibility because as long as $N_{\mathrm{safe}}>0$ the MPC algorithm retains feasibility.

    Having $N_{\mathrm{safe}}=r$ implies that from the current state $x_0$ we can reach the safe set $\Ss$ in $r$ time steps, i.e., $x_0 \in \Bs_r(\Ss)$.
    By assumption~\eqref{eq:condition_rf_pcmpc} we know that $\Bs_r(\Ss) \subseteq \bigcup_{j=r+1}^N \Bs_j(\Ss)$.
    Putting it all together we can infer that $x_0 \in \bigcup_{j=r+1}^N \Bs_j(\Ss)$.
    This means that from $x_0$ it is possible to reach $\Ss$ in $j$ time steps, for some value $j \in [r+1, N]$.
    Given that, we know that the OCP~\eqref{eq:parallel_constr_mpc} must be feasible using $x_0$ as initial state for $p=j$, implying that $N_{\mathrm{safe}}$ will be increased (at least) to $j-1$ in line ~\ref{line:N_safe_parallel} of Alg.~\ref{alg:parallel_mpc}.
\end{proof}

%
The reason for the weaker requirements of PC-MPC w.r.t. RC-MPC and HTC-MPC is hidden in their working principles. 
Indeed, RC-MPC features two safety constraints in its formulation (a hard receding constraint and a soft terminal constraint). 
Both of them must be satisfied to slide the receding constraint forward and ensure RF. 
Therefore, the system must be able to reach $\Ss$ at the terminal node starting from $\Ss$ at the receding-constraint node.
HTC-MPC is somewhat more similar to PC-MPC because they both work with a single safety constraint per OCP. 
However, while HTC-MPC only tries to satisfy it at the terminal node, PC-MPC tries it on all nodes of the horizon, which reasonably leads it to weaker requirements.

\subsection{Summary of safe sets and MPC formulations}
\label{ssec:sets_summary}
We have presented five types of multi-step CISs, which generalize the concept of CIS, as depicted in Fig.~\ref{fig:safe_sets}.
To summarize, we report here the different safe-set conditions.
Classic CISs must satisfy the following condition:
\begin{eqs}{eq:control_invariance}
\Ss \subseteq \Bs_1(\Ss).
\end{eqs}
$K$-step CISs must satisfy:
\begin{eqs}{eq:strong_n_step_ci}
\Ss \subseteq \Bs_K(\Ss).
\end{eqs}
Up-to-$N$-step CISs must satisfy:
\begin{eqs}{eq:up_to_n_step_ci} 
\exists \, K \in [1, N]: \quad \Ss \subseteq \Bs_K(\Ss).
\end{eqs}
Within-$N$-step CISs must satisfy:
\begin{eqs}{eq:weak_n_step_ci} 
\Ss \subseteq \bigcup_{j=1}^N \Bs_j(\Ss).
\end{eqs}
Back-and-forth-$N$-step CISs must satisfy:
\begin{equation} \label{eq:baf_n_step_ci}
    \exists \, k \in [0, N-1]: \quad \Bs_k(\Ss) \subseteq \Bs_N(\Ss).
\end{equation}
Finally, back-and-forth-within-$N$-step CISs must satisfy:
\begin{equation} \label{eq:condition_rf_pcmpc2}
    \exists \, r \in [0, N-1]: \quad \Bs_r(\Ss) \subseteq \bigcup_{j=r+1}^N \Bs_j(\Ss).
\end{equation}
\review{These sets are related to BRS, and Appendix~\ref{app:back_reach} discusses their relationship.
Fig.~\ref{fig:safe_sets} shows the relationships between these set classes, highlighting that most of them can be seen as relaxations of the previous ones.}
We have shown how three MPC formulations from the literature, namely RC-MPC, HTC-MPC, and PC-MPC, can guarantee RF or safety using these novel multi-step CISs.

Unfortunately, the exact computation of any of these multi-step CISs remains as challenging as determining a standard CIS. 
In practice, however, numerical procedures for approximating CISs are more likely to produce a set that belongs to one of the newly introduced classes, rather than a CIS. 
Therefore, as empirically suggested by our results, these approaches tend to be safe with higher probability compared to a conventional MPC scheme, even though, in practice, the assumptions underlying Theorems~\ref{th:recursive_feasibility}, \ref{th:within_recursive_feasibility} and \ref{th:bf_recursive_feasibility} cannot be guaranteed.
\section{Safe Task Abortion} 
\label{sec:task_abortion}
Since we do not have any theoretical guarantee on the set $\Ss$, in practice we cannot ensure RF of any of the MPC schemes discussed above. Thus, we rely on the safe-abort strategy proposed in~\cite{Lunardi2024} to safely stop the robot in case of unfeasibility.
To ensure safety we rely on the following three assumptions.
\begin{assumption} \label{ass:conservative_set}
    The safe set $\mathcal{S}$ is a conservative approximation of the set $\Bs_M(\mathcal{E})$ : 
    \begin{equation}
        \Ss \subseteq \Bs_M(\mathcal{E}).
    \end{equation}
    Note that this assumption does not require that $\Ss$ satisfy any type of multi-step control invariance.
\end{assumption}
\begin{assumption} \label{ass:computational_units}
\review{To use HT-MPC and RC-MPC we assume the availability of two computational units, denoted as unit A and unit B. 
To use PC-MPC we assume the availability of N+1 computational units.}
\end{assumption}
\review{In this paper, a computational unit denotes an independent execution resource, such as a processor core or computational thread, capable of running one control-related computation in parallel with the others.}
\begin{assumption} \label{ass:ocp_time}
We can solve the following OCP for any $x_{\mathrm{init}} \in \Ss$, in $L + 1$ time steps (with $L \in [0, N-1]$):
\begin{equation}
    \begin{aligned}
        \minimize_{\{x_i\}_0^{M+1},\{u_i\}_0^{M}}  &\quad  1 \\
        \st & \quad \eqref{eq:mpc_initial_conditions}, \eqref{eq:mpc_dynamics}, \eqref{eq:mpc_bound_constraints} \\
            &\quad x_{M} = x_{M+1}.
    \end{aligned}
    \label{eq:safe_abort_ocp}
\end{equation}
\end{assumption}
OCP~\eqref{eq:safe_abort_ocp} can be employed to compute a feasible trajectory that steers the system from $x_{\mathrm{init}}$ to an equilibrium state. \review{Since only constraint satisfaction is relevant for the abort procedure, we use $1$ as a dummy cost function}.
In the following we discuss how the safe-abort procedure is implemented with the three MPC schemes discussed in Section~\ref{sec:safe_mpc}.

\subsection{Safe Abort with HTC-MPC}
Let us consider the HTC-MPC formulation given by problem~\eqref{eq:htc_mpc}, and assume that at iteration $k$ the MPC problem becomes unfeasible. 
In this scenario, the task can be safely aborted by following these steps:
\begin{enumerate}
    \item unit A employs the MPC solution computed at iteration $k-1$ while attempting to solve the MPC problem~\eqref{eq:htc_mpc} for the current state of the robot;
    \item simultaneously, unit B solves OCP~\eqref{eq:safe_abort_ocp}, taking $x_{N|k-1}$ as initial state; 
    \item if the system reaches $x_{N|k-1}$ (because the MPC problem resulted to be unfeasible for $N$ consecutive iterations), the solution of OCP~\eqref{eq:safe_abort_ocp} is followed to safely bring the system to an equilibrium state. 
\end{enumerate}
This strategy allows us to reach a safe equilibrium state, from which we can try to solve again the MPC. 
We have noticed that sometimes this approach leads to completing the task after the abort procedure, increasing the success rate.   
This strategy ensures safety, as formalized in the following Lemma.
\begin{lemma}
    Under Assumptions~\ref{ass:conservative_set} to~\ref{ass:ocp_time}, the HTC-MPC (outlined in Alg.~\ref{alg:terminal}) with safe task abortion (described above) guarantees that constraints are never violated.
\end{lemma}
\begin{proof}
This proof follows directly from the stated assumptions. 
OCP~\eqref{eq:safe_abort_ocp} is always feasible because, by Assumption~\ref{ass:conservative_set}, any state within $\Ss$ can reach an equilibrium in at most $M$ time steps.
Assumption~\ref{ass:ocp_time} (with $L=N-1$) ensures that dedicating a computational unit to solving OCP~\eqref{eq:safe_abort_ocp} provides a solution before reaching the terminal state $x_{N|k-1}$ of the last feasible MPC problem.
Once $x_{N|k-1}$ is reached, following the solution of OCP~\eqref{eq:safe_abort_ocp} safely drives the system to an equilibrium state, where it can remain indefinitely without violating any constraints.
\end{proof}
The most critical assumption is probably Assumption~\ref{ass:ocp_time}, which requires the MPC horizon $N$ to be sufficiently long, and that $M$ is not excessively large, to provide adequate time to solve OCP~\eqref{eq:safe_abort_ocp}.
However, contrary to~\cite{Lunardi2024}, we suggest to choose $M$ so that OCP~\eqref{eq:safe_abort_ocp} is not too computationally demanding. 
Of course this implies a trade-off because choosing a small value of $M$ would result in a small safe set $\mathcal{S}$, leading in turn to poor performance, so we must choose $M$ as large as possible, while satisfying the computation time constraint.

\subsection{Safe Abort with RC-MPC and PC-MPC}
\label{ssec:abort_receding}
Let us assume that we are using the RC-MPC formulation, and that the receding constraint on node $r$ has reached time step $L$. 
At this point we would need to trigger the task-abortion strategy because we have only $L+1$ time steps to solve OCP~\eqref{eq:safe_abort_ocp}. 
However, for ensuring RF of RC-MPC under the assumptions of Theorem~\ref{th:recursive_feasibility}, we need to allow $r$ to reach node 0.
This would imply solving the safe abort OCP in zero time, which is impossible.
Therefore, to ensure consistency with the safe abort procedure, while maintaining RF, we propose an extended version of RC-MPC.
This new version (summarized in Alg.~\ref{alg:receding_w_abort}) does not alter the computed trajectory after $r$ has reached node $L$, unless $r$ can be moved forward.

Suppose that at time step $t$ the receding constraint $r$ has reached $L$.
At this point we must use Unit B to start solving the safe-abort OCP~\eqref{eq:safe_abort_ocp} using the predicted state \review{$x^*_{L+1|t}$} as initial state.
In parallel, unit A can be used to keep trying to solve the RC-MPC formulation from the current state.
If the solution of RC-MPC satisfies the terminal constraint, we can resume normal operation, setting $r=N-1$.
Otherwise, we use the solution computed at time step $t$ to control the system, until we reach $x^*_{r|t}$, and then we follow the solution of the safe-abort OCP to reach an equilibrium state.

\begin{algorithm}[t]
\caption{Receding-Constraint MPC with Safe Abort}
\small
\begin{algorithmic}[1]
\Require 
Number of time steps $T$,
Initial state $x_0$,
\mbox{Initial guess $\{x^g_i\}_0^{N}, \{u^g_i\}_0^{N-1}$},
OCP~\eqref{eq:rc_mpc}, 
Safe-abort OCP~\eqref{eq:safe_abort_ocp},
Safe-abort trigger time~$L$ 
    \State $r \leftarrow N$  \Comment{Receding constraint index}
    \For{$t = 0 \rightarrow T-1$}
        \If{$r = 0$} \Comment{Abort task}
            \State $\textsc{FollowSafeAbortTrajectory()}$
            \State $r \leftarrow N$ \Comment{Reset}
        \EndIf
        \State $\{x_i^*\}_0^{N}, \{u_i^*\}_0^{N-1} \leftarrow \textsc{OCP}(r, x_t, \{x^g_i\}_0^{N}, \{u^g_i\}_0^{N-1})$
        \If{$r \le L$ \textbf{and} $x_N^* \notin \Ss$}
        		\State $\{x_i^*\}_0^{N}, \{u_i^*\}_0^{N-1} \leftarrow \{x^g_i\}_0^{N}, \{u^g_i\}_0^{N-1}$
        \EndIf
        \For{$j = r+1 \rightarrow N$} \Comment{Search last state in $\Ss$}
            \If{$x_j^* \in \Ss$}
                \State $r \leftarrow j$ \Comment{Move receding constraint forward}
            \EndIf
        \EndFor
        \State $r \gets r - 1$  \Comment{Recede constraint}
        \If{$r = L$} \Comment{Trigger Safe Abort}
            \State $\textsc{SolveSafeAbortOCP}(\review{x_{L+1|t}^*})$
        \EndIf
        \State $x_{t+1} \leftarrow f(x_t,u_0^*)$ \Comment{Simulate dynamics}
        \review{\State $\{x^g_i\}_0^{N}, \{u^g_i\}_0^{N-1} \leftarrow \textsc{WarmStart}(\{x_i^*\}_1^{N}, \{u_i^*\}_1^{N-1})$}
    \EndFor
\end{algorithmic}
\label{alg:receding_w_abort}
\end{algorithm}

The same strategy outlined above for RC-MPC can also be used for PC-MPC.
When the receding constraint node $r$ has reached the value $L$ we start the safe-abort computation.
At the same time we stop using the new trajectories computed by PC-MPC, unless they satisfy the safety constraint on a node after $L$. 
We use instead the last solution computed when $r=L+1$.

\subsection{Solving the Safe Abort OCP}
Clearly the safe abort OCP should be solved as efficiently as possible to guarantee good performance of the controller.
One possible strategy is to pre-compute a warm-start for OCP~\eqref{eq:safe_abort_ocp}, before $r$ reaches 0. 
Although the exact state of the system at that moment is unknown, the trajectory predicted by the MPC can be used as initial guess.
If this warm-start proves to be insufficient, RC-MPC can be modified to ensure that the pre-computed safe-abort trajectory begins precisely at the state where the task abortion is triggered.
To accomplish this, the receding constraint can be adjusted from $x_{r} \in \Ss$ to the more conservative $x_{r} = x_{r+1}^g$, effectively forcing the predicted state within $\Ss$ to remain unchanged across the MPC cycles.
However, this strategy is expected to reduce overall performance: indeed, we have tested this solution in simulation and found that it performs poorly. 
The adopted solution in our work was to rely on the Real-Time Iteration scheme~\cite{diehl2005nominal} for solving OCP~\eqref{eq:safe_abort_ocp} (further details are provided in Section~\ref{ssec:safe_abort_computation}). 
In this case, Assumption~\ref{ass:computational_units} is no longer required: when the safe abort procedure is triggered, unit A swaps the MPC problem with OCP~\eqref{eq:safe_abort_ocp} and solves it in one time step, without any involvement of unit B.

\section{Safe Set Computation}
\label{sec:safe_set_computation}
This section provides some details on how the safe set used in the MPC formulations is computed.
%
Among the different methods mentioned in Section~\ref{ssec:mpc_ci} for approximating CISs, we chose VBOC~\cite{LaRocca2023}. 
Indeed, VBOC is an efficient algorithm for fully-actuated systems (like robot manipulators) since it can sample states directly on the border of the safe set. 
In the original work, VBOC is used to approximate the \emph{viability kernel}, which is assumed to be equivalent to the $\infty$-step BRS of the equilibria $\Bs_\infty(\mathcal{E})$.\review{We note that the two notions are not identical in full generality: $\Bs_\infty(\mathcal{E})$ imposes asymptotic convergence to $\mathcal{E}$, whereas viability only requires indefinite constraint satisfaction.}
VBOC finds states exactly on the set's boundary $\partial \Bs_\infty(\mathcal{E})$ through the following OCP:
\begin{equation} 
    \begin{aligned}
        \maximize_{\{ x_i\}_{0}^{M+1}, \{ u_i\}_{0}^{M}} &\quad  a^\T  x_0, \\
        \st &\quad  x_{i+1} =  f( x_i,  u_i),  \\
        &\quad  x_i \in \Xs, \,  u_i \in \Us(x_i), \quad \forall \, i=0,\dots,M, \\
        &\quad  H  x_0 =  h, \\
        &\quad  x_M =  x_{M+1},
    \end{aligned}
\label{eq:viab_boundary_ocp}
\end{equation}
where the quantities $a$,  $H$, and $h$ are defined as:
\begin{equation}
    \label{eq:compute_a_S_s}
    \begin{aligned}
     a = \begin{bmatrix} 0 \\  d \end{bmatrix},
    \quad
     H = \begin{bmatrix}  I & 0 \\ 0 &  I -  d  d^\T \end{bmatrix},
    \quad
     h = \begin{bmatrix}   q_{\text{init}} \\ 0 \end{bmatrix}.
    \end{aligned}
\end{equation}
The main insight of OCP~\eqref{eq:viab_boundary_ocp} is to maximize the velocity norm of the initial state $\norm{ v_0}$ by fixing the initial configuration to $q_{\text{init}}$ and the direction of the initial velocity in joint space to $d$, chosen as uniformly random values. Even though the shape of $\Bs_\infty(\mathcal{E})$ is not known a priori, uniform sampling of $ q_{\mathrm{init}}$ and $d$ allows a uniform coverage of the input space of our set representation, i.e., the space of joint angles and joint velocity directions. The constraint $ H  x_0 =  h$ ensures $ x_0^* \in \partial \Bs_\infty(\mathcal{E})$ (all the proofs can be found in~\cite{LaRocca2023}).

\review{Under the assumption adopted in the original VBOC formulation that the relevant viability boundary coincides with the boundary of $\Bs_\infty(\mathcal{E})$}, the horizon $M$ must be sufficiently long: starting from a reasonable initial guess, problem~\eqref{eq:viab_boundary_ocp} is repeated with an increasing $M$, until the optimal cost $ a^\T  x_0^*$ converges.

The first state obtained by~\eqref{eq:viab_boundary_ocp} is always on $\partial \Bs_\infty(\mathcal{E})$, and in the paper the authors have proposed an algorithm to check if the successive states $\{ x_i\}_1^M$ are also on $\partial \Bs_\infty(\mathcal{E})$. 

We have slightly modified the algorithm proposed in~\cite{LaRocca2023} to better perform in our setting. Indeed, the original implementation did not consider the presence of Cartesian obstacles in the environment. In addition, in our work we are interested in $M$-step BRS $\Bs_M(\mathcal{E})$, which are subsets of the $\infty$-step BRS $\Bs_\infty(\mathcal{E})$, therefore the iterative procedure to increase the horizon $M$ is no longer needed for our purpose.

\begin{algorithm}[t]
    \caption{Modified version of the VBOC algorithm} 
    \small
    \begin{algorithmic}[1]
    
    \Require 
    Number of trajectories $K$,
    Time horizon $M$,
    Lower and upper bounds of the joint positions $q_{\mathrm{min}}, q_{\mathrm{max}}$,
    OCP~\eqref{eq:viab_boundary_ocp}
    
    \State $\mathcal{D} \leftarrow [\,]$
    \State $k \leftarrow 0$
    \Repeat   
        \State $q \leftarrow \textsc{RandomUniform}(q_{\mathrm{min}}, q_{\mathrm{max}})$
        \If{$\textsc{IsCollisionFree}(q)$}
            \State $d \leftarrow \textsc{RandomNormal}(0, 1) $
            \State $d \leftarrow d / \norm{d}$
            \State $a, H, h \leftarrow$~\eqref{eq:compute_a_S_s}
            \State $\{x_i^*\}_{0}^{M+1}, \{u_i^*\}_{0}^{M} \leftarrow \textsc{OCP}(a, H, h)$
            \State $\textbf{insert } x_0^* \textbf{ in } \mathcal{D}$
            \State $k \leftarrow k + 1$
        \EndIf
    \Until{$k = K$}
    \State $\mathcal{D} \leftarrow \textsc{RebalanceVelocityNorm}(\mathcal{D}) $
    \State \Return $\mathcal{D}$
    \end{algorithmic}
    \label{alg:vboc}
\end{algorithm}
The modified version of the algorithm is shown in Alg.~\ref{alg:vboc}. Given the presence of Cartesian obstacles inside the environment, after the sampling of a joint configuration we must check if it is in collision (Line 5). In case the configuration is feasible, the VBOC problem can be solved (Line 10). The process is repeated until we find $K$ states on $\partial \Bs_M(\mathcal{E})$, collected in the dataset $\mathcal{D}$. 

Algorithm~\ref{alg:vboc} needs more samples than the original formulation, since it only uses the first state computed by the OCP, even though other states from that trajectory could be on $\partial \Bs_M(\mathcal{E})$. However, this comes with the great benefit of having uniformly distributed data, which would not be the case using the original algorithm. Moreover, the computational cost of our approach is rather close to the one of the previous implementation: indeed, each OCP takes less time to be solved, since the horizon $M$ is fixed and smaller than the one used in~\cite{LaRocca2023}. 

As in the original work, the dataset $\mathcal{D}$ is used to train a NN $\phi(\cdot)$, which takes as input the joint configuration and the velocity direction, and outputs the maximum velocity norm. $\phi(\cdot)$ is used to express the approximated safe set $\Ss$, exploiting it in an MPC formulation to achieve safety. 
We have noticed that the NN tends to exhibit larger prediction errors for values of $\norm{v}$ close to zero. This behavior is attributed to the non-uniform distribution of $\norm{v}$ in the dataset $\mathcal{D}$, which is heavily biased toward higher velocity values. To address this issue, we rebalance the dataset $\mathcal{D}$ (Line 13 in Alg.~\ref{alg:vboc}) by copying data points so that the data frequency is approximately uniform across the spectrum of $\norm{v}$.

\section{Software Implementation}
\label{sec:software}
This section discusses the software implementation of the MPC methods. In particular, we focus on how to integrate the nonlinear dynamics of the robot manipulator and the approximated CIS inside a unique MPC formulation for Alg.~\ref{alg:receding_w_abort} and Alg.~\ref{alg:terminal},~\ref{alg:parallel_mpc} with safe task abortion. We have used \textsc{CasADi} \cite{Andersson2019} for the symbolic computation of the dynamics, costs, and constraints, and \textsc{Acados}~\cite{Verschueren2019} to solve the OCPs and integrate the dynamics. The whole framework has been developed using Python. 

\subsection{Robot dynamics}
To integrate the dynamics of robotic manipulators using \textsc{CasADi} symbols, we have used the \textsc{Adam} library, which implements a set of algorithms for calculating rigid-body dynamics for robotic systems starting from Universal Robotic Description Format (URDF) files. In our implementation, the joint accelerations $\ddot q$ are considered as control input $u$, thus the system dynamics is linear (i.e., it is a double integrator). Then, we have used the Recursive Newton-Euler Algorithm~\cite{featherstone2014rigid} to compute the joint torques and impose their bounds as nonlinear constraints in~\eqref{eq:mpc_bound_constraints}:
\begin{equation}
    \tau_\mathrm{min} \le M(q) u + C(q, \dot q) + g(q) \le \tau_\mathrm{max},
    \label{eq:torque_limits}
\end{equation}
where $M(\cdot)$ is the mass matrix, $C(\cdot)$ contains the Coriolis and centrifugal effects, $g(\cdot)$ is the gravitational term, while $\tau_\mathrm{min}, \tau_\mathrm{max}$ are the joint torque bounds. This is an important extension with respect to~\cite{Lunardi2024,Fontanari2025}, where only a simple dynamics (triple pendulum system) with hand-written equations of motion has been considered.

\subsection{Collision modeling}
Unlike in~\cite{Lunardi2024,Fontanari2025}, where only joint constraints were considered, this work accounts for Cartesian obstacles and self-collision avoidance of the robotic arm.  
We modeled the robot links and the obstacles inside the environment using simple geometric shapes: spheres, capsules, and planes.
This allows us to check collisions with a low computational effort, while ensuring the differentiability of the collision constraints, which is required by the MPC solver.
Actually, distance between capsules is not differentiable everywhere because its computation contains up to 6 evaluations of the min and max functions.
However, \textsc{Acados}~\cite{Verschueren2019} can deal with this kind of functions that are differentiable almost everywhere.
%

\subsection{Mismatch between MPC state and NN input}
Set membership to $\Ss$ is verified with the following constraint~\cite{Lunardi2024,Fontanari2025}:
\begin{equation} 
    \label{eq:nn_constraint}
    (1-\alpha) \phi(x) - ||\dot{q}|| \ge 0,     
\end{equation}
where $\phi(\cdot)$ is a NN computing an upper bound on the joint velocity norm as described in Section~\ref{sec:safe_set_computation}, and $\alpha \in [0, 1]$ is a safety margin that we introduced to satisfy Assumption~\ref{ass:conservative_set}. We have used \textsc{L4CasADi}~\cite{salzmann2023learning} for integrating the PyTorch~\cite{pytorch} neural model inside \textsc{Acados}. 

The VBOC algorithm~\cite{LaRocca2023}, used for the computation of the safe set, suffers from the well-known curse of dimensionality. Indeed, the complexity of the algorithm scales exponentially with the dimension of the state,\review{ which limits the practical applicability of VBOC for high-dimensional systems such as full robotic manipulators, whose state dimension is typically around 12.}

Given $n_j \in \mathbb{N}$ as the number of joints of the manipulator, a possible solution to mitigate this issue is to derive the safe set considering only a subset of $m \in \mathbb{N}$ joints, with $m < n_j$. The remaining $n_j - m$ joints are considered static in the configuration $\bar q$ and neglected by the VBOC algorithm. Still, the space occupied by all the links of the robot in the environment, even the fixed ones, is taken into account. Then, assuming without loss of generality that the $m$ considered joints are the first ones, we can check set membership to $\mathcal{S}$ with the following constraints:
\begin{equation}
    \label{eq:nn_constraint_mis}
    \begin{aligned}
        (1-\alpha) \phi((q[:m], \dot q[:m])) - ||\dot q[:m]|| &\ge 0 \\
        q[m:] &= \bar q \\
        \dot q[m:] &= 0.
    \end{aligned}
\end{equation}
This is an additional contribution with respect to~\cite{Lunardi2024,Fontanari2025}, that allows us to validate our approach on a full 6-DOF robot manipulator (details are provided in Section~\ref{sec:experiments}).

\subsection{Real-Time Iteration scheme}
\label{ssec:rti_scheme}
MPC requires to solve optimization problems inside fast control loops, which is hard for high-dimensional systems. Indeed, the converging solution of OCP~\eqref{eq:mpc} cannot be found online at each sampling time. Our MPC implementation relies on the Real-Time Iteration (RTI) scheme \cite{diehl2005nominal}, which boils down to performing a single iteration of Newton's method, without line search.
Therefore, after each solver iteration we need to check if constraints are satisfied.
In particular, also the torque limits can be violated, that in our case is expressed as a nonlinear constraint.
For this reason, for every problem, we perform a check on the state bounds for all the running nodes. Also, we verify the satisfaction of the obstacle avoidance constraints. Then, the MPC step is accepted if the previous conditions are satisfied, otherwise the previous solution is used. In case the joint torques do not satisfy the bounds~\eqref{eq:torque_limits}, $\tau$ is saturated and the corresponding acceleration is computed through the forward dynamics, followed by an update of the state. The NN constraint~\eqref{eq:nn_constraint} instead is not checked after each OCP solution: empirically we have noticed that demanding the exact satisfaction of~\eqref{eq:nn_constraint} results in a lot of unnecessary safe aborts.
\review{Notice that the feasibility checks and fallback mechanisms described above are applied only to the three proposed methods, namely HTC-MPC, RC-MPC, and PC-MPC.}

\subsection{Safe-Abort Computation}
\label{ssec:safe_abort_computation}
The safe abort strategy described in Section~\ref{sec:task_abortion} requires the solution of OCP~\eqref{eq:safe_abort_ocp}. The best scenario for RC-MPC and PC-MPC is that the safe abort OCP can be solved in one time step. For this reason, we have applied the RTI scheme also for this case, \review{thus $L = 0$}. Indeed, we noticed through simulations that solving OCP~\eqref{eq:safe_abort_ocp} using the RTI scheme leads to the same number of safely aborted tasks as using Sequential Quadratic Programming until convergence, while having a computation time lower than one time step. This is a contribution with respect to~\cite{Lunardi2024,Fontanari2025}, where no solutions were explored to solve the OCP~\eqref{eq:safe_abort_ocp} in one time step.

\section{Simulation Results}
\label{sec:simulations}


\subsection{Simulation Setup}
To evaluate the effectiveness of the analyzed safe MPC formulations, we simulated them together with three classic MPC formulations of increasing complexity:
\begin{itemize}
    \item Naive: it has no terminal constraint, i.e.,  problem~\eqref{eq:mpc} with $\mathcal{X}_N = \Xs$. This is the baseline for all the experiments. 
    \item Zerovel: it enforces an equilibrium terminal state ($x_N \in \mathcal{E}$), which is the most conservative approach to get RF.
    \item STC-MPC: it introduces a soft terminal constraint set $\mathcal{X}_N = \Ss$ with a penalty weight of $w_s=10^4$, as an attempt towards RF (Section~\ref{ssec:recursive_feasibility}).
\end{itemize} 
For the simulations, we have considered the Z1 joint manipulator from Unitree\textregistered. The horizon for  VBOC~\eqref{eq:viab_boundary_ocp} and safe abort~\eqref{eq:safe_abort_ocp} has been fixed to $M = 45$. \review{Note that the baseline methods do not incorporate the safe-abort mechanism: Naive and Zerovel do not have access to the safe set, and although STC-MPC enforces feasibility of the solution, it does not provide any safety guarantees.}

\subsection{Setpoint regulation}
\label{ssec:simu_setpoint}
\begin{figure}[tb]
    \centering
    \includegraphics[width=0.49\columnwidth, trim = 440 200 400 100, clip]{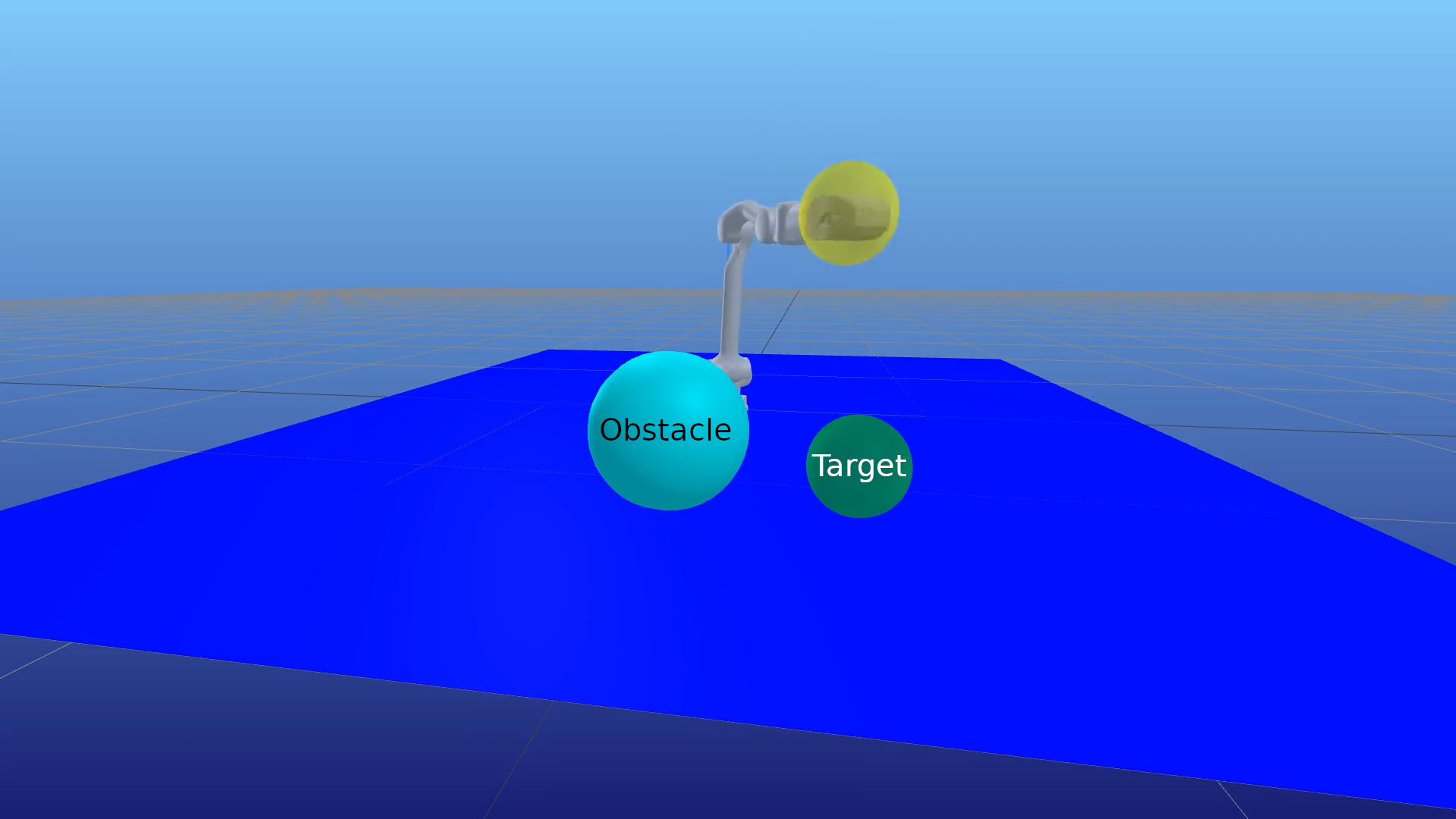}
    \includegraphics[width=0.49\columnwidth, trim = 440 200 400 100, clip]{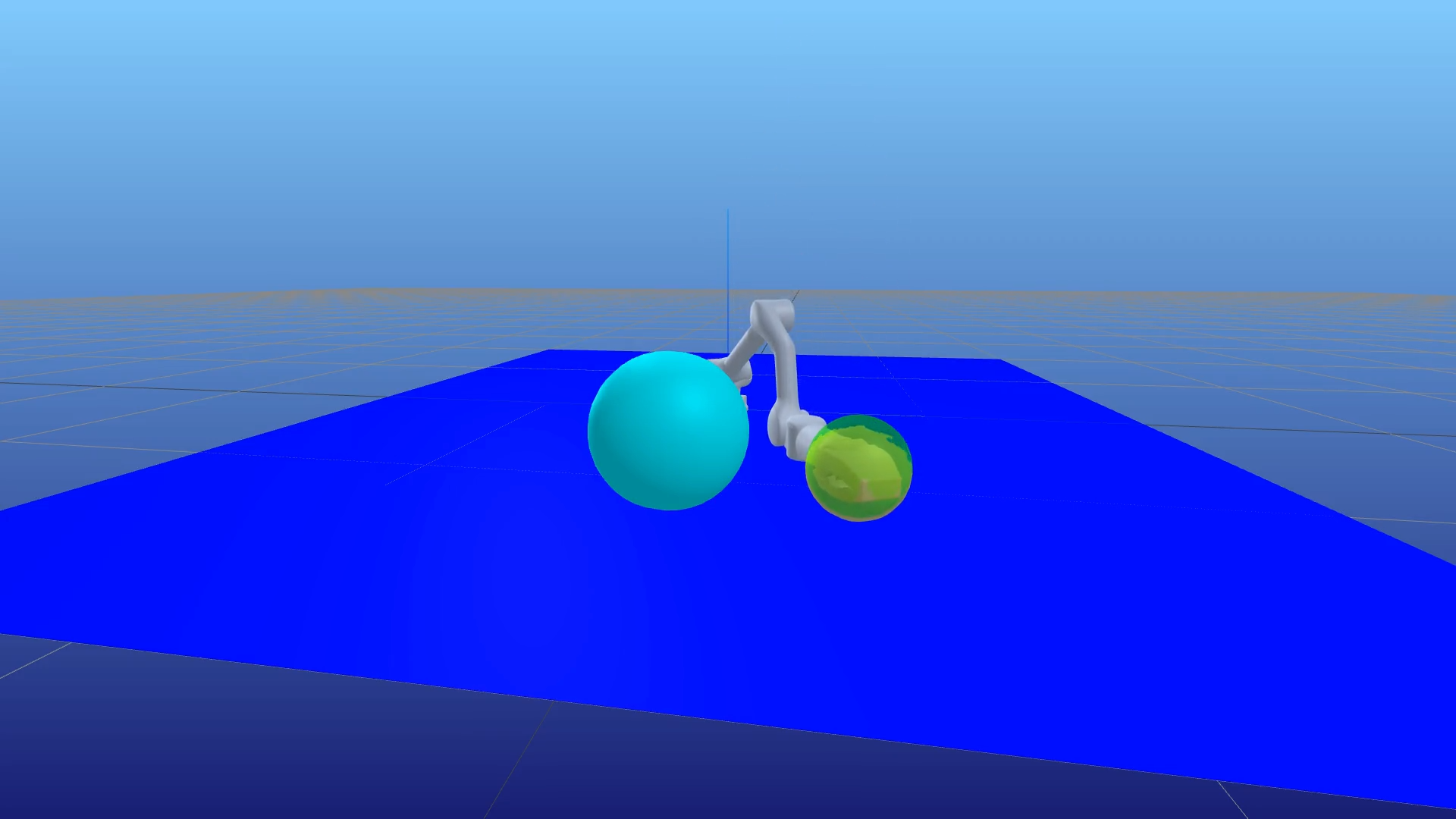}
    \caption{Simulation setup for the setpoint regulation task. 
    The robot starts from a random initial configuration (left) and must reach a target with the EE (right), avoiding collisions with the ball and the ground.}
    \label{fig:env_setup_setpoint}
\end{figure}
The first task is a setpoint regulation problem (depicted in Fig.~\ref{fig:env_setup_setpoint}), where the End-Effector (EE) of the robot has to reach a target position (in green) close to a ball (cyan) and ground (blue) obstacles, avoiding collisions. 
The EE bounding volume is defined by a sphere obstacle (faded yellow).
Given $p_e, p_e^\mathrm{ref} \in \Rv{3}$ as the current and reference EE positions, we have used a least-squares function as running cost, which penalizes deviations from $p_e^\mathrm{ref}$ and control efforts:
\begin{equation}
    \begin{aligned}
        \ell(x, u) &= \norm{p_e(x)-p_e^\mathrm{ref}}_Q^2 + \norm{u}_R^2, \\
        Q &= 500 \, I, \quad R = 5 \cdot 10^{-3} \, I,
    \end{aligned}
    \label{eq:weight_values}
\end{equation}
\review{Notice that we deliberately refrain from constraining the end-effector velocity, because the goal of this paper is to assess whether our proposed MPC schemes can enforce safety on a given controller, even when that controller alone might result in unsafe behavior.}
We have run $400$ simulations for each MPC formulation, starting from the same $400$ random joint positions $q_0$ with $\dot{q}_0 = 0$. \review{First, we check the configuration $q_0$ to being not in collision with obstacles; then, we try to compute an OCP solution to warm-start the subsequent MPC controller. Since at beginning the optimal safety condition for each safe-set based controller is the same, i.e. satisfaction of the safe set constraint at the end
of the control horizon, the only controllers that we used to generate the initial conditions and the initial warm-start trajectories are Naive, Zerovel, and HTC. } The MPC time step is $\mathrm{d}t = \SI{5}{\milli\second}$, while the total simulation lasts $n = 800$ time steps. 
The robot is commanded on the first 4 joints, while the remaining two are fixed, thus $x \in \Rv{8},\, u \in \Rv{4}$. The NN has 2 hidden layers with 256 neurons and GELU activation functions to guarantee differentiability of~\eqref{eq:nn_constraint}. The training was performed on a dataset of 600k points, generated using the VBOC algorithm on a 32-cores machine, requiring approximately 10 hours.
\begin{figure*}
    \centering
    \subfloat[\label{fig:setpoint_hor_fail}]{\includegraphics[width=\columnwidth]{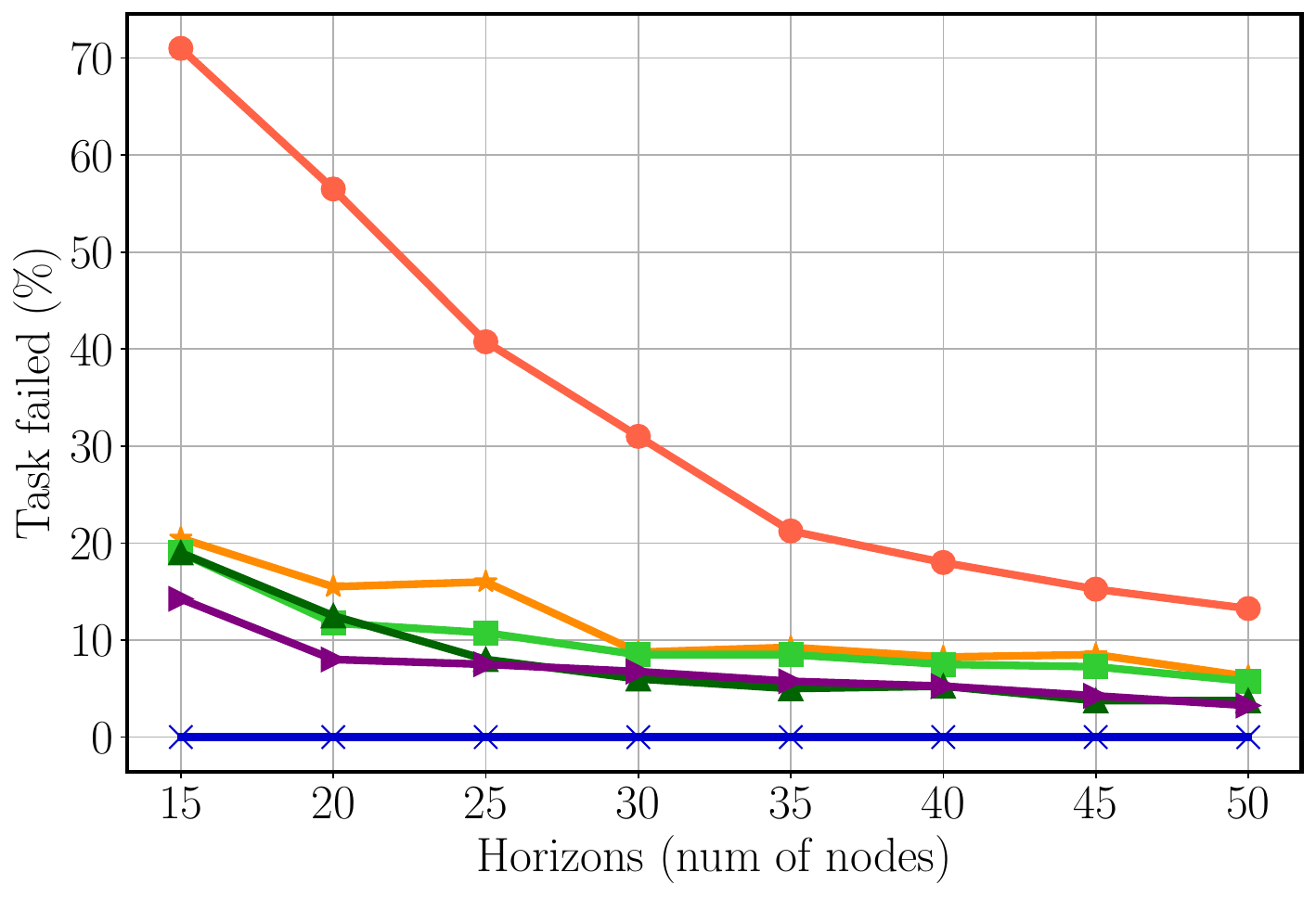}} \,
    \subfloat[\label{fig:setpoint_hor_score}]{\includegraphics[width=\columnwidth]{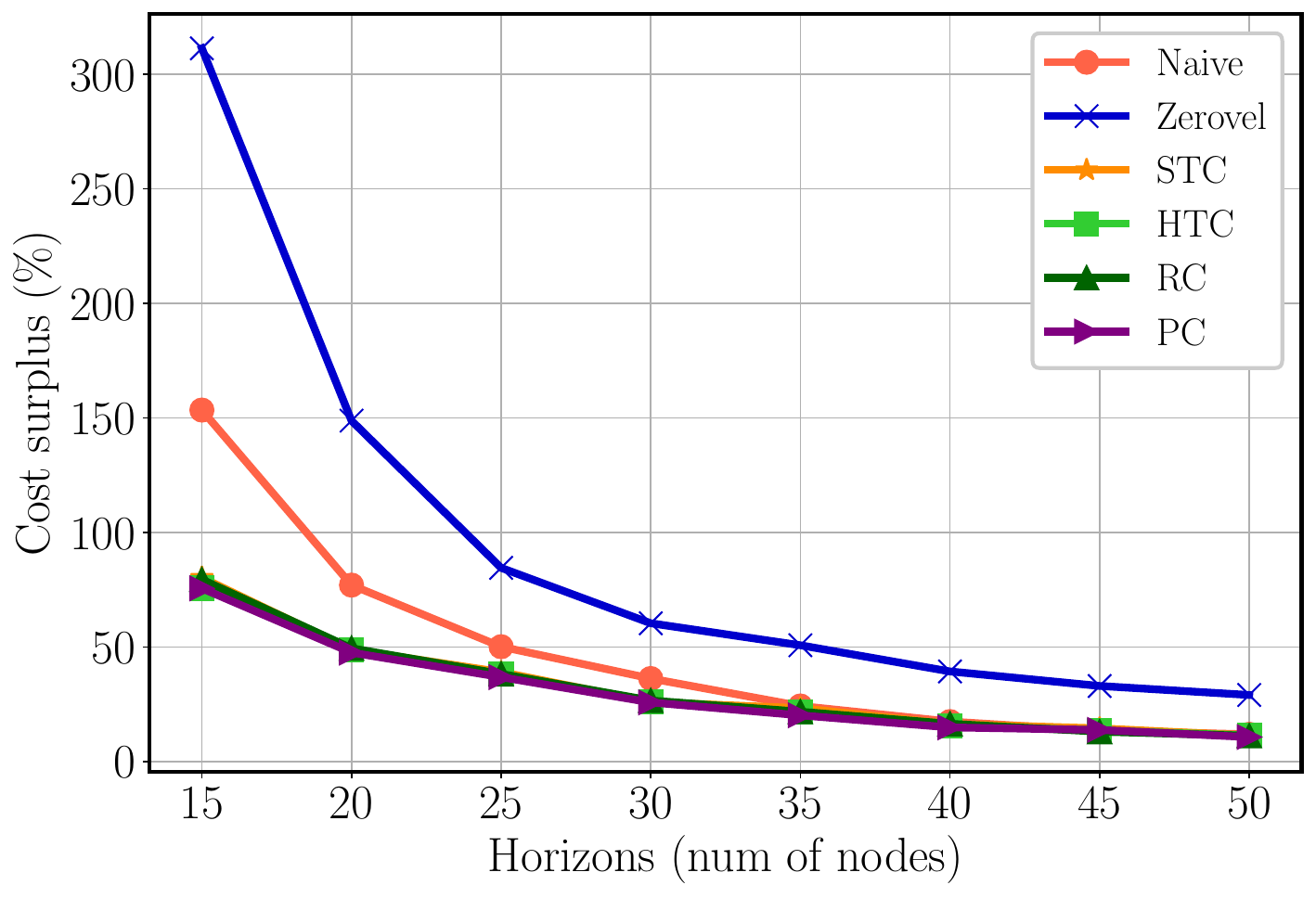}} \\
    \subfloat[\label{fig:setpoint_alpha_fail}]{\includegraphics[width=\columnwidth]{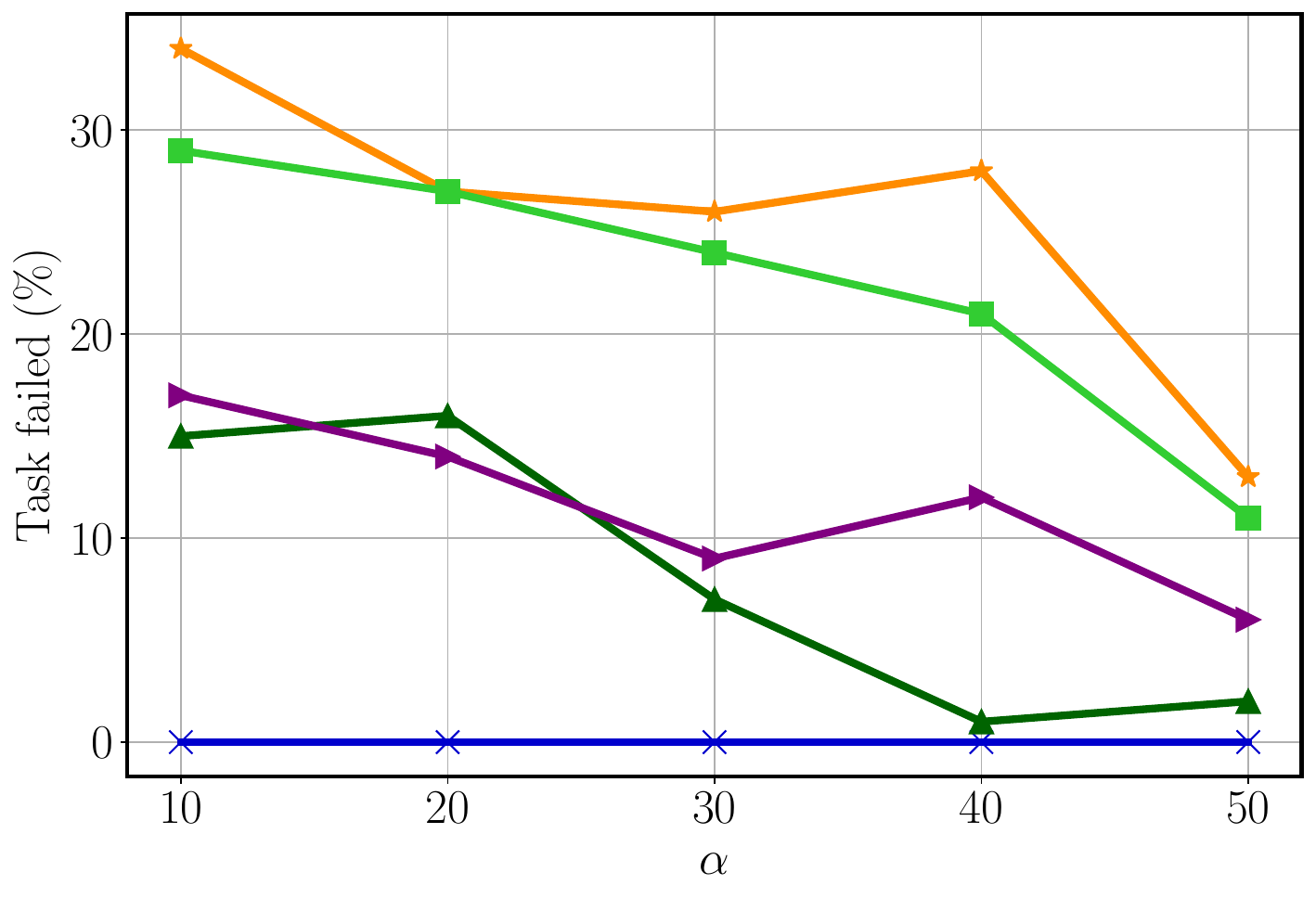}} \,
    \subfloat[\label{fig:setpoint_alpha_score}]{\includegraphics[width=\columnwidth]{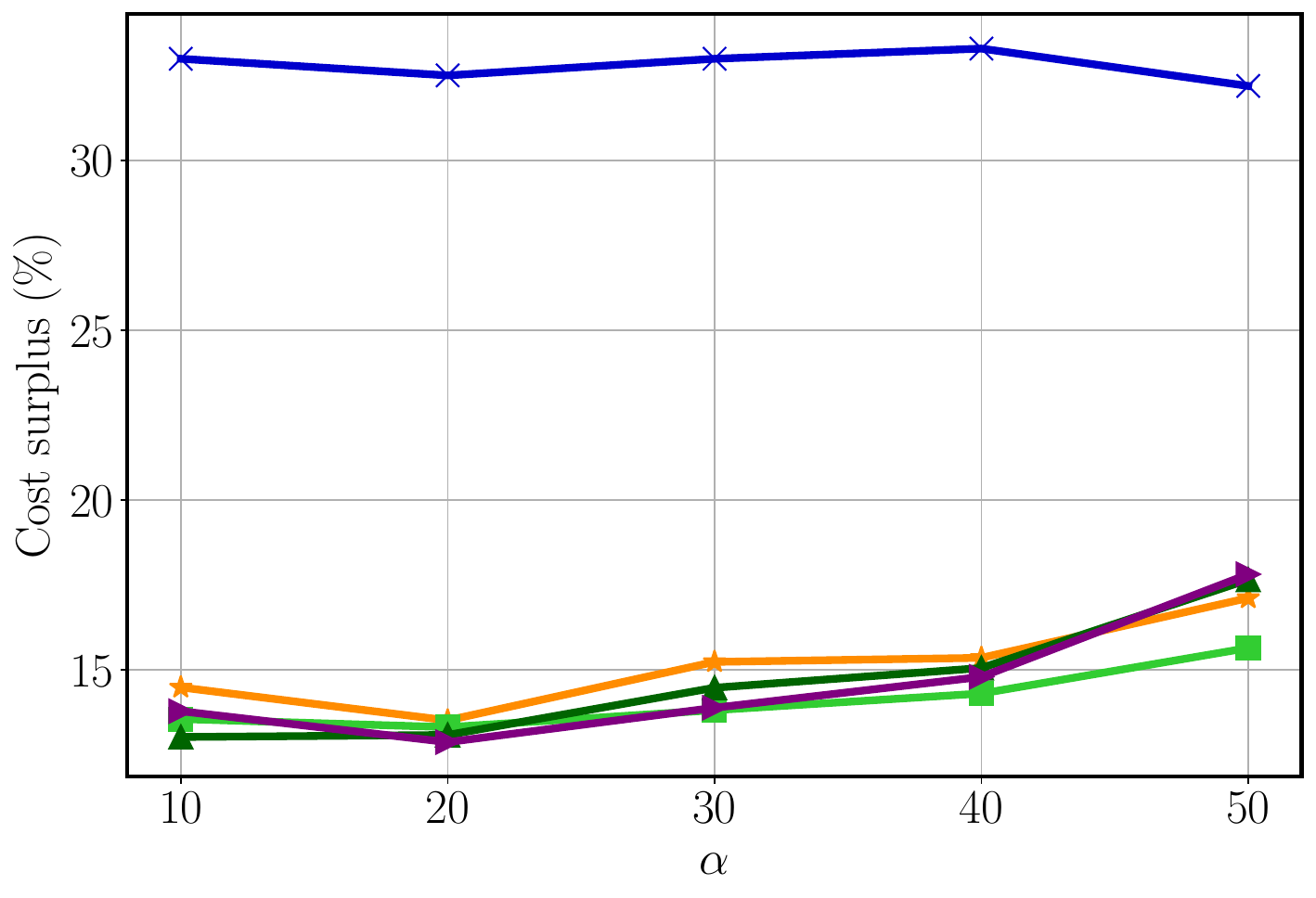}} 
    \caption{Simulation results of the different MPC controllers for the setpoint regulation task, considering different values of horizons and NN safety margins.}
    \label{fig:setpoint_results}
\end{figure*}

We know from~\cite{Diehl2017book} that the horizon length $N$ has a strong influence on RF and performance of MPC, thus we decided to analyze how the results change with different values of the horizon. Figure~\ref{fig:setpoint_hor_fail} shows how the number of failed tasks changes with respect to the horizon: a task is considered as failed if the robot hits the joint bounds or the EE collides with any obstacle. Naive is the least safe method, with almost $70\%$ of the tests failed with $N = 15$. Indeed, we know that RF is practically impossible for short horizons without using an appropriate terminal constraint. Excluding Zerovel, which never fails, the other formulations are less sensitive to the horizon. In addition, for short horizons ($N\le 30$), HTC-MPC, RC-MPC, and PC-MPC fail less than the STC-MPC approach.

To evaluate the performance of the MPC formulations, we focus our attention on the cost function~\eqref{eq:mpc_cost}: the lower the value of $\ell$, the better the tracking performances. Given $J$ as the total cost of a simulation, thus:
\begin{equation}
    J = \sum_{i=0}^{n-1} \ell_i(x_i, u_i) + \ell_n(x_n),
\end{equation}
we define the \emph{cost-surplus} as:
\begin{equation}
    \label{eq:surplus}
    \sigma_\mathrm{mpc} = \left(\review{\frac{1}{\nu}}\sum_{i = 1}^{\nu} \frac{J_\mathrm{mpc,i}}{J^*_i} - 1\right) \cdot 100,
\end{equation}
where the subscript $\mathrm{mpc}$ indicates the formulation used, and $\nu$ is the number of tasks in which no controller failed (if any controller fails a task, that task is excluded from the computation of the cost-surplus). The optimal cost $J^*$ is computed by solving an OCP on the full horizon, thus $N = n$. This OCP results in a high number of optimization variables, thus a good warm-start and a robust solver are necessary: for this purpose, we have used the trajectories $\{x_i\}_0^n,\, \{u_i\}_0^{n-1}$ obtained during the MPC simulations, taking the ones performing the best in terms of cost, among the different formulations for each initial condition; for the solver, we chose \textsc{Ipopt}~\cite{Wachter2006}. The higher $\sigma$, the worse the tracking performance.

Figure~\ref{fig:setpoint_hor_score} compares the cost-surplus, among the $\nu$ tests, of each formulation against the horizon. For Zerovel, the cost-surplus rapidly increases for smaller horizons, since the predicted velocity $\dot{q}$ must reach zero in a smaller time. Instead, for the other formulations the cost-surplus increases slowly as expected, due to less conservative terminal/receding constraints. Surprisingly, the cost-surplus of the novel formulations are slightly lower than Naive: empirically, we noticed that the EE with the Naive controller often overshoots the target, while the other formulations do so less often since they include a safety constraint on the terminal and/or running node.  

The choice of the safety margin $\alpha$ is another critical parameter for Alg.~\ref{alg:receding_w_abort} and Alg.~\ref{alg:terminal},~\ref{alg:parallel_mpc} with safe task abortion. Indeed, Assumption~\ref{ass:conservative_set} may be not valid if $\alpha$ is too small. The larger the safety margin, the less the false positives of the NN. 
We evaluated the variations on the number of failed tasks and the cost-surplus by choosing different values of $\alpha$. The MPC horizon has been fixed to $N=45$. Notice that Naive is excluded from this analysis because it is independent of $\alpha$, while Zerovel is maintained for comparison. In Figure~\ref{fig:setpoint_alpha_fail}, the number of failed tasks decreases with higher values of $\alpha$ for the methods with safe abort procedure. In particular, RC-MPC and PC-MPC are the safest approaches, with RC-MPC failing less than $2\%$ of the times for $\alpha \ge 30\%$. 

Larger $\alpha$ implies a smaller safe set, thus the cost-surplus increases as a drawback. This behavior can be noticed in Fig.~\ref{fig:setpoint_alpha_score}: for the constraint formulations, the cost-surplus inflates only by a small quantity, from $\approx 15\%$ to $\approx 18\%$. These values are much lower with respect to Zerovel, that has a cost-surplus always larger than $30\%$. The reason for this increase is that with higher $\alpha$ the safe MPC approaches activate more frequently the safe abort procedure, thus the system avoids failures but the robot stops typically at a large distance from the target before restarting the task, resulting in larger tracking cost.

\begin{figure}[tb]
    \centering
    \includegraphics[width=\columnwidth]{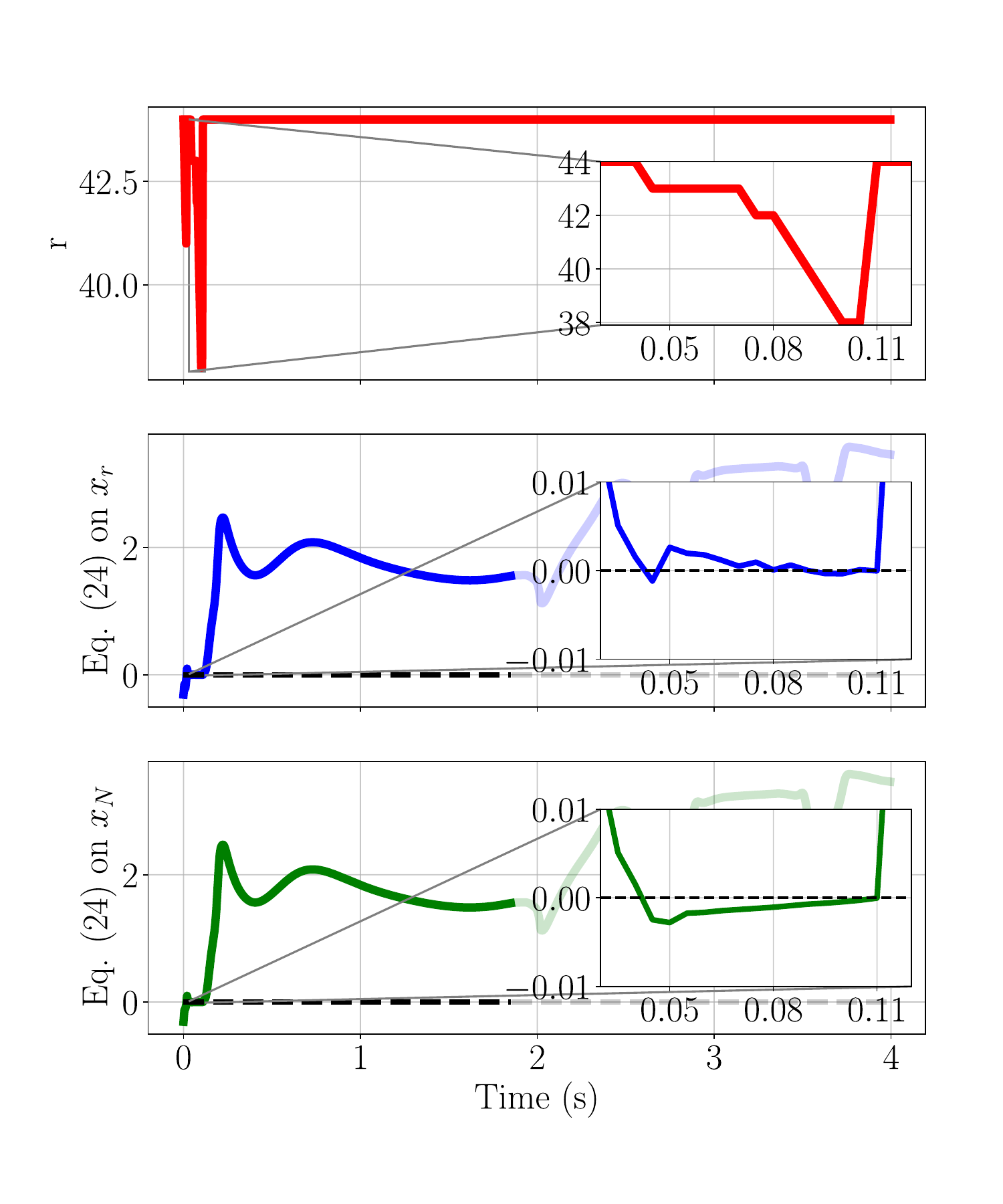}
    \caption{Values of $r$ and NN outputs (\review{left-hand side} of \ref{eq:nn_constraint})
    evaluated at $x_r^*$ and $x_N^*$, in one particular simulation, with horizon $N=45$ and safety margin $\alpha = 10\%$.}
    \label{fig:r_vs_nn_out}
\end{figure}
The variable $r$ in RC-MPC plays a crucial role in guaranteeing RF under the assumption that the set $\Ss$ is up-to-$N$-step control invariant. Figure~\ref{fig:r_vs_nn_out} illustrates the output of the NN, defined by~\eqref{eq:nn_constraint}, evaluated at the predicted states $x_r^*$ and $x_N^*$. Throughout the simulation, $r$ remains nearly constant at $44$ (that is $N-1$) except for a brief drop at the beginning of the test.
A closer inspection at the interval between $t = \SI{0.03}{\second}$ and $t = \SI{0.12}{\second}$ reveals a reduction in $r$ when the NN output at $x_N^*$ becomes negative. As outlined in Algorithm~\ref{alg:receding_w_abort}, this triggers the receding mechanism, where $r$ decreased until a state $x_j^* \in \Ss$ is found (Lines 9-11). At $t = \SI{0.11}{\second}$, the NN output for the terminal state becomes positive again, and $r$ returns to $44$.
Theoretically, the NN output evaluated at the receding state should remain strictly positive. However, a careful examination of the middle plot shows that it becomes slightly negative at certain points. As discussed in Section~\ref{ssec:rti_scheme}, this occurs because the RTI optimization scheme does not guarantee constraint satisfaction and we decided not to check the satisfaction of constraint~\eqref{eq:nn_constraint} at every MPC iteration because this would lead to premature safe aborts.

\subsection{Computation Time}

\begin{table}[tb]
    \begin{center}
        \caption{99-percentile computation times for the MPC using the RTI scheme.}
        \label{table:comp_times}
        {\renewcommand\arraystretch{1.1}
        \begin{tabular}{lccccc}
            \toprule
            & \sc{Naive} & \sc{Zerovel} & \sc{STC} & \sc{HTC} & \sc{RC} \\
            \midrule
            \sc{RTI} [$\SI{}{\milli\second}$] & 4.68 & 4.18 & 7.76 & 4.15 & 4.78 \\
            \bottomrule
        \end{tabular}
        }
    \end{center}
\end{table}
Table~\ref{table:comp_times} reports the 99th percentile computation times of the MPC iterations for each method. These results refer to the 400 simulations of the setpoint regulation task presented in Section~\ref{ssec:simu_setpoint}. This metric provides a conservative estimate that includes many of the worst-case (i.e., slowest) iterations. For all the formulations except STC-MPC, the 99th percentile remains below the duration of the time step, allowing the MPC to be recomputed at each time step in real-time. In the current implementation, PC-MPC solves N subproblems~\eqref{eq:parallel_constr_mpc} sequentially rather than in parallel. Thus, the computation times are not reported, but we expect them to be comparable to those of HTC-MPC, since each PC-MPC subproblem involves only a single constraint~\eqref{eq:nn_constraint} on a single node. 

For what concerns the time required for the safe abort control computation, the times are almost always under the time step duration (5 ms) of the MPC in the simulation campaign of Section~\ref{ssec:simu_setpoint}, as can be seen in Table \ref{table:safe_backup_times}, allowing an efficient computation of the controls to drive the system to a safe configuration.

\begin{table}[tb]
    \begin{center}
        \caption{\review{RTI safe abort time statistics, for simulation with horizon $N=20$, $\alpha = 40\%$ on 423 samples}.}
        \label{table:safe_backup_times}
        {\renewcommand\arraystretch{1.1}
        \begin{tabular}{lc}
            \toprule
            Mean $[\SI{}{\milli\second}]$                & \; 1.71  \\
            STD $[\SI{}{\milli\second}]$                 & \; 0.48  \\
            99-quantile $[\SI{}{\milli\second}]$         & \; 2.59  \\
            Times $<$ 5 $\SI{}{\milli\second}$ [\%] & \; 99.76 \\
            \bottomrule
        \end{tabular}
        }
    \end{center}
\end{table}

\subsection{Robustness evaluation}
\label{ssec:simu_disturbance}
\begin{figure}[tb]
    \centering
    \includegraphics[width=0.49\columnwidth, trim = 400 200 600 50, clip]{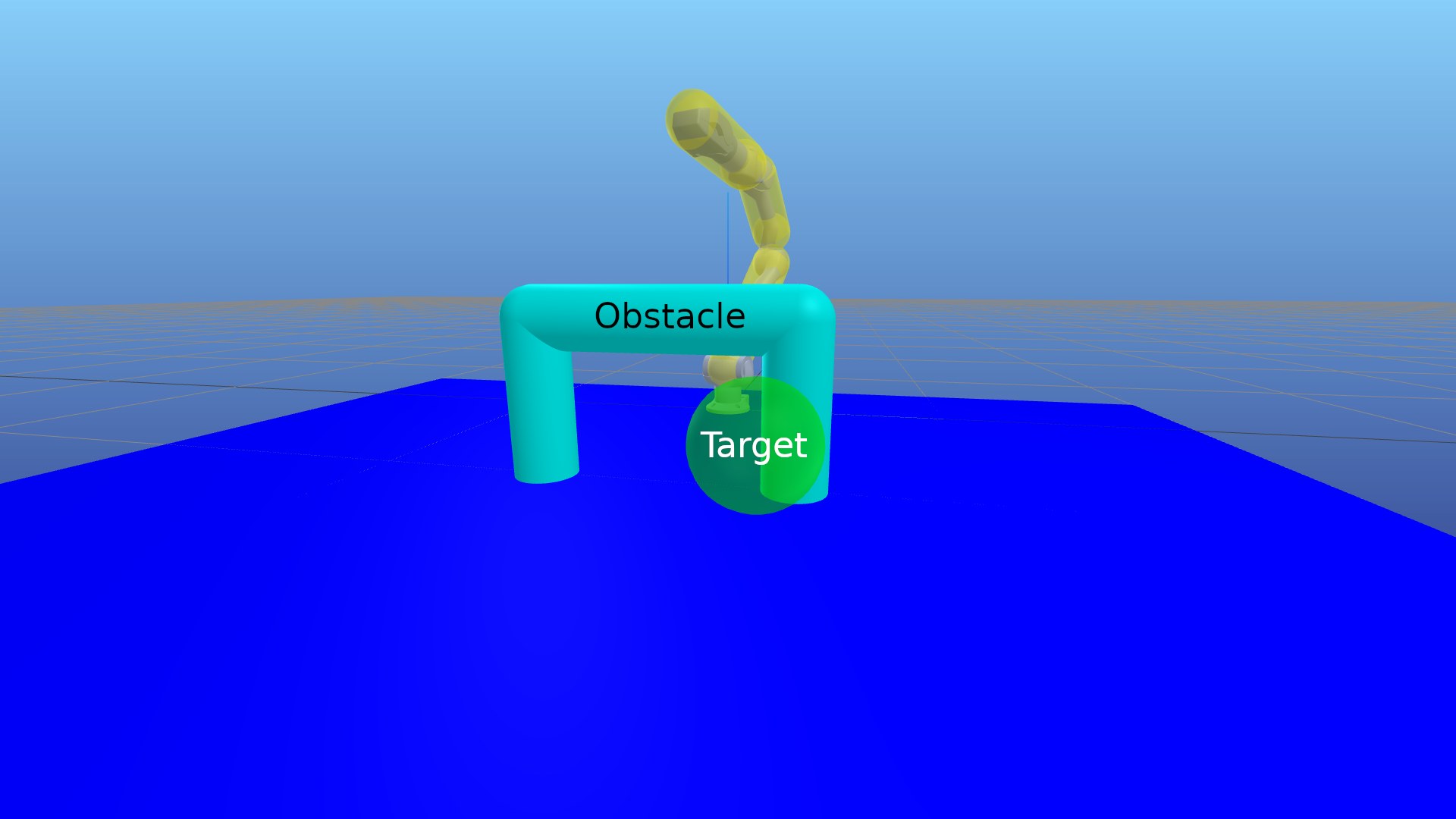}
    \includegraphics[width=0.49\columnwidth, trim = 400 200 600 50, clip]{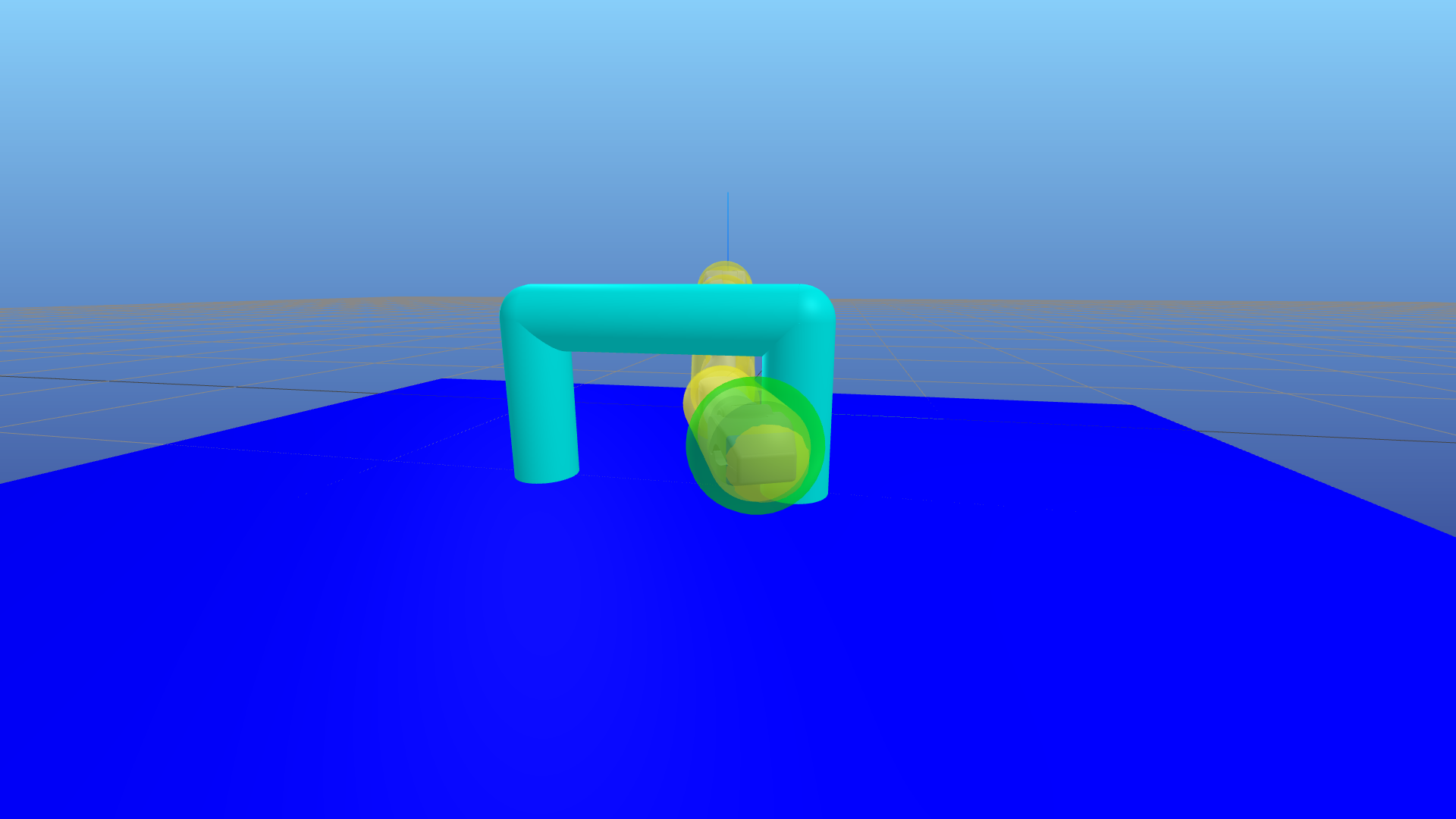}
    \caption{Simulation setup to evaluate disturbance rejection capability in the setpoint task.  The robot starts from a fixed initial configuration (left) and must reach the EE's target (right), avoiding collisions with the ground and the window-shaped obstacle.}
    \label{fig:env_setup_disturbace}
\end{figure}

\begin{figure*}
    \centering
    \subfloat[\label{fig:dist_noise_fail}]{\includegraphics[width=\columnwidth]{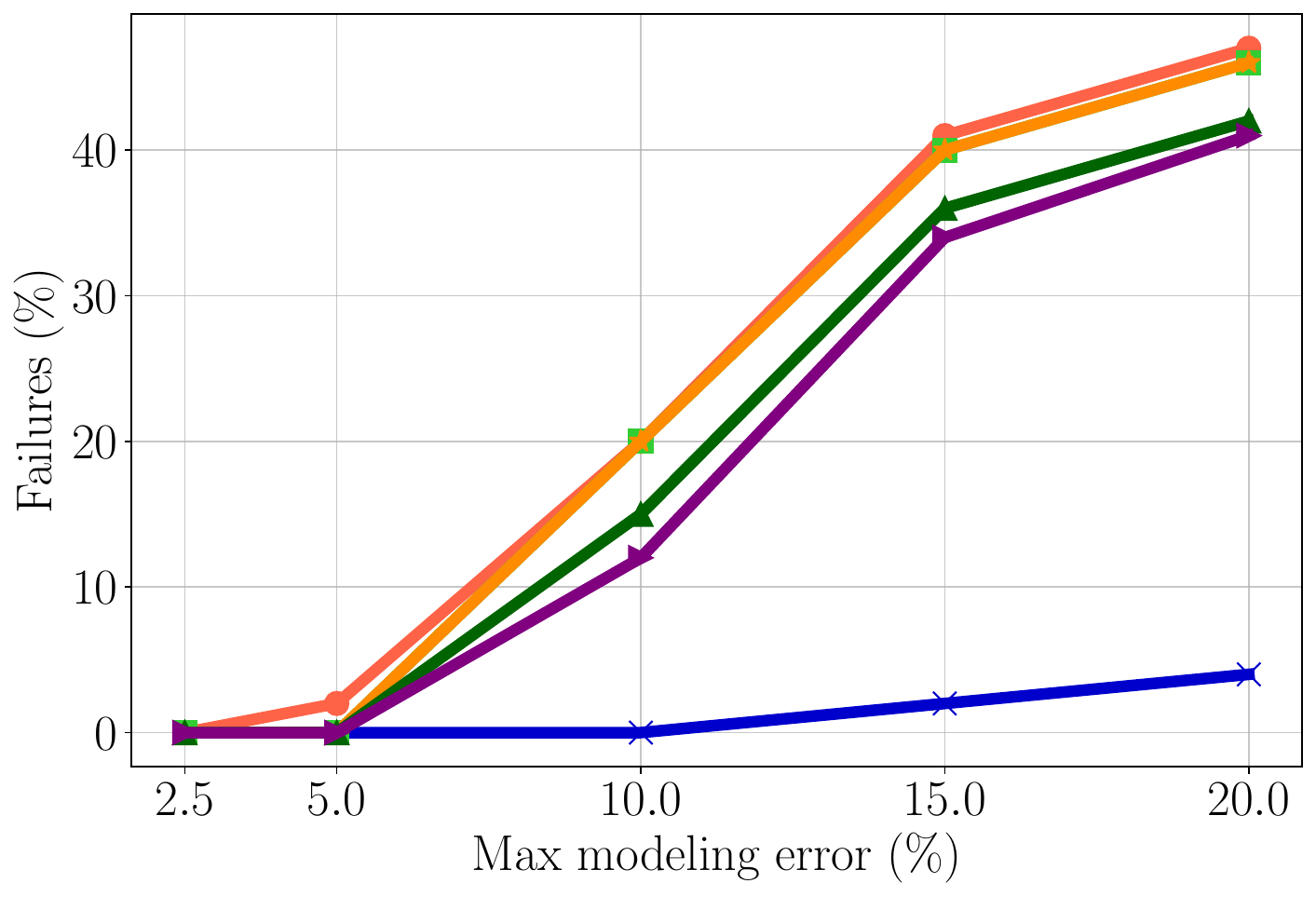}} \,
    \subfloat[\label{fig:dist_noise_score}]{\includegraphics[width=\columnwidth]{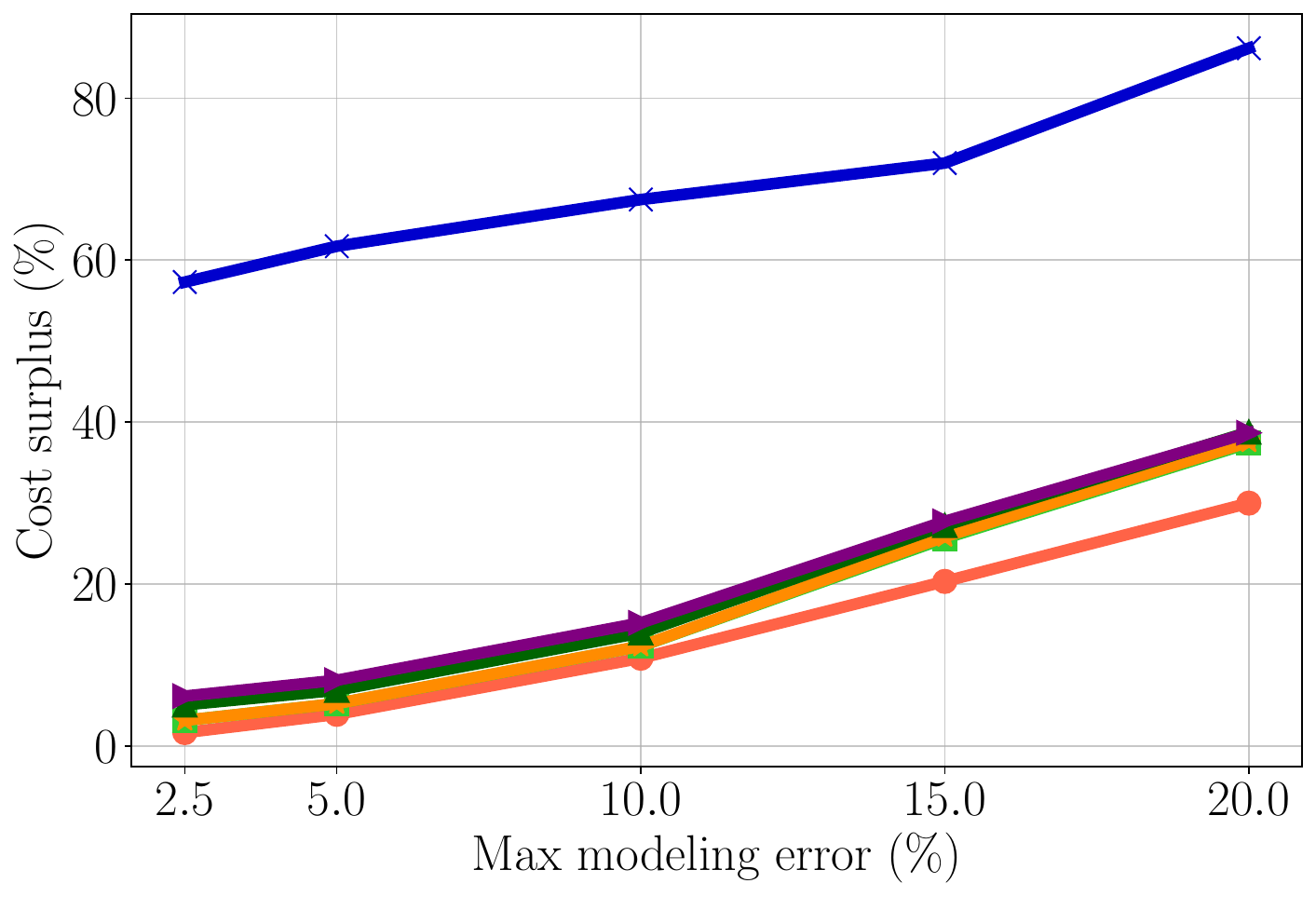}} \\
    \subfloat[\label{fig:dist_margin_fail}]{\includegraphics[width=\columnwidth]{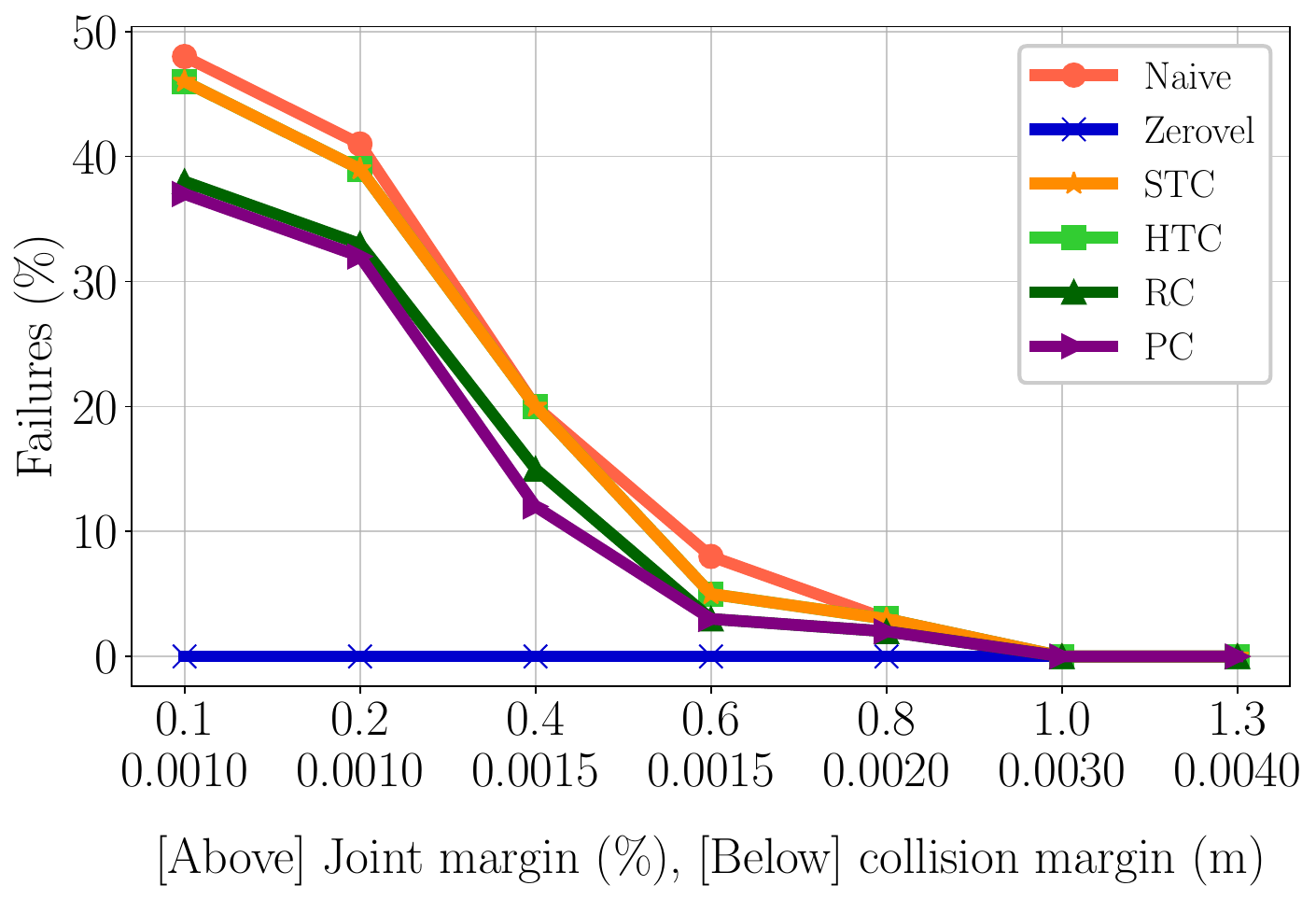}} \,
    \subfloat[\label{fig:dist_margin_score}]{\includegraphics[width=\columnwidth]{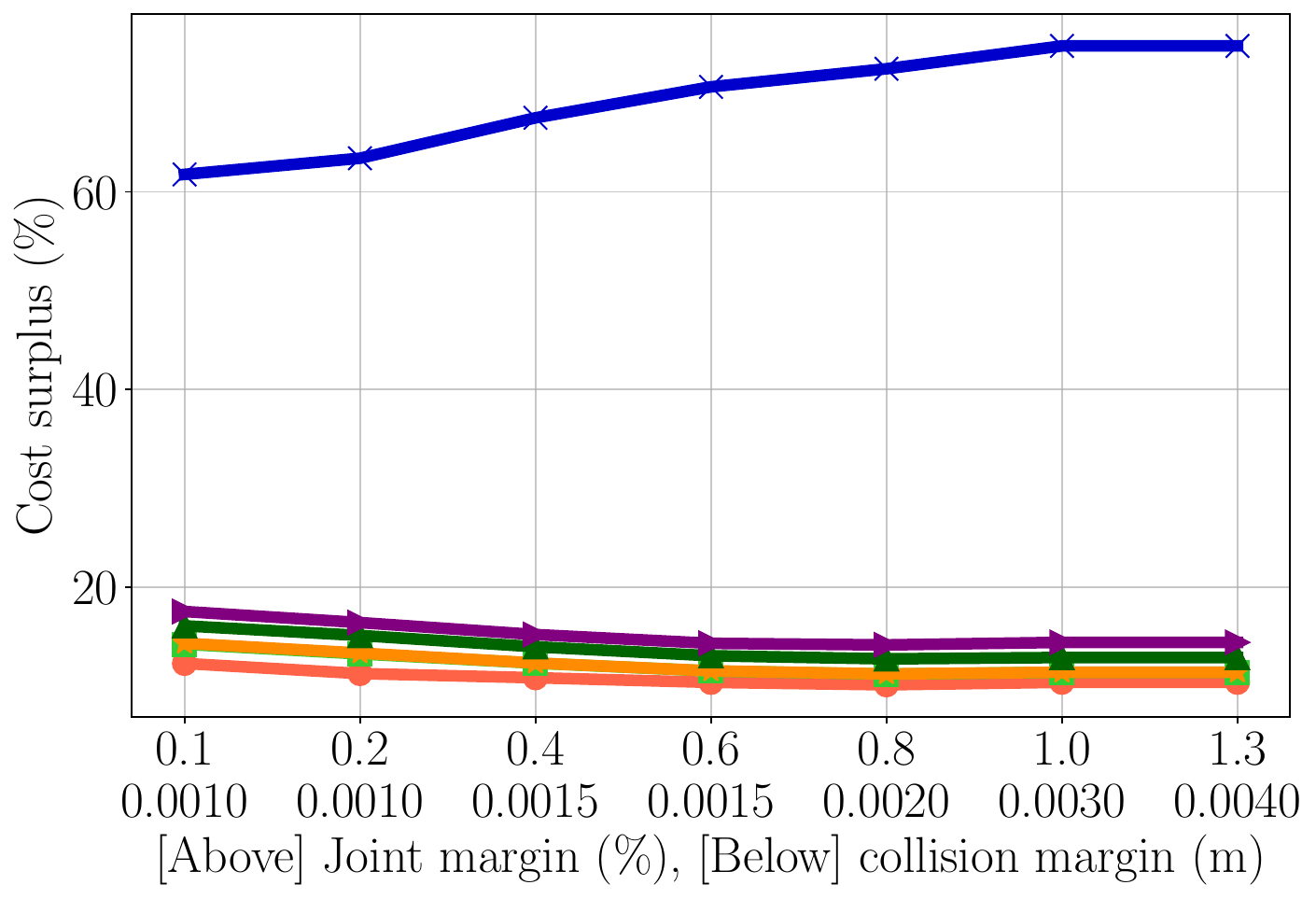}} 
    \caption{Simulation results with modeling errors, considering different values of parameter noise and joint/collision margins. The horizon and NN safety margin $\alpha$ for these tests are respectively $N = 25$ and $\alpha = 20 \%$.}
    \label{fig:disturbace_results}
\end{figure*}

Using the same setpoint regulation task described in Section~\ref{ssec:simu_setpoint}, we evaluate the robustness of the safe MPC methods to modeling errors. The setup for these simulations is shown in Fig.~\ref{fig:env_setup_disturbace}. In this scenario, the EE must reach the green target avoiding collisions with the plane (blue) and a window-shaped obstacle (cyan), modeled by three capsules. Unlike the previous setup, collisions are checked over the full bounding volume of the robot, represented using capsules (faded yellow). Here, five joints of the robot are controlled ($x \in \Rv{10}, u \in \Rv{5}$), while the wrist joint is fixed. 
The MPC time step is $\mathrm{d}t = \SI{10}{\milli\second}$, while the total simulation lasts $n = 800$ time steps as before. 
A new NN, using the same architecture as before, is trained on a dataset of 1.2 million points, generated in approximately 72 hours.

To simulate modeling errors, we randomly perturb the inertial parameters of the robot links by a maximum factor of $\delta_\lambda$, using uniform sampling. 
Since the control methods discussed in this paper do not directly account for modeling errors, we need to introduce some hand-tuned safety margins in the state constraints (i.e., joint limits and obstacle dimensions). Typically, such margins should be introduced in the controller constraints. However, this would imply re-training the NN used in the constraint~\eqref{eq:nn_constraint} for every margin. To avoid this computational overhead, we decided to keep the controller constraints unaltered, while changing instead the joint limits and obstacle dimensions in the simulation environment. Given a user-defined margin $\delta_j$ and $x_\Delta \triangleq |x_\mathrm{max} - x_\mathrm{min}|$, the enlarged bounds are:
\begin{equation}
    \tilde{x}_\mathrm{min} = x_\mathrm{min} - x_\Delta \frac{\delta_j}{100}, \quad \tilde{x}_\mathrm{max} = x_\mathrm{max} + x_\Delta \frac{\delta_j}{100}.
\end{equation}
Similarly, we shrink the obstacles in the simulator by adding a positive margin $\delta_o$ to the distance computation. 

Figure~\ref{fig:dist_noise_fail} reports the number of failed tasks as a function of the noise level $\delta_\lambda$, using joint and obstacle margins of $\delta_j = 0.4\%$ and $\delta_o =  \SI{0.0015}{\meter}$. For each value of $\delta_\lambda$, we perform 100 simulations, sampling each time different inertial parameters, and using the same initial robot configuration shown in Fig.~\ref{fig:env_setup_disturbace}. As expected, the number of failures increases with higher modeling errors. Naive is the worst controller, while the safe MPC formulations, particularly RC-MPC and PC-MPC, demonstrate significantly higher robustness, excluding the over-conservative Zerovel controller. Figure~\ref{fig:dist_noise_score} shows the cost-surplus $\sigma$ across different values of modeling errors. Here, $J^*$ from equation~\eqref{eq:surplus} is computed as the minimum cost $J$ among all simulations and controllers for each value of $\delta_\lambda$. The high safety of Zerovel comes at a price of a high cost-surplus, reflecting poor tracking performance. In contrast, the safe controllers maintain values of $\sigma$ comparable to Naive. 

A similar analysis is presented in Fig.~\ref{fig:dist_margin_fail} and~\ref{fig:dist_margin_score}, where we evaluate the effect of different joint and obstacle margins $\delta_j$ and $\delta_o$, while keeping the modeling errors fixed at $\delta_\lambda = 10\%$. The larger the margins, the lower the number of \review{failed tasks}. With the exception of Zerovel, the safe MPC formulations consistently result in fewer failures compared to Naive. In particular, for small safety margins, RC-MPC and PC-MPC exhibit $\approx 10 \%$ fewer failures than Naive and the terminal constraint controllers. The cost-surplus in Fig.~\ref{fig:dist_margin_score} follows the same trend as in Fig.~\ref{fig:dist_noise_score}, highlighting the trade-off between safety and performance.
\review{Note that the modeling error margins used in this experiment are not meant to provide an exhaustive characterization of the uncertainty of the model. Rather, they were manually selected as representative mismatch levels to illustrate the qualitative evolution of failures and cost surplus for the considered MPC schemes, while keeping the figure readable.}

\subsection{\review{Set statistics}}
\label{ssec:set_statistics}
\review{To empirically support our hypothesis that it is easier to find sets with weaker membership conditions compared to CIS, and that the use of an approximate CIS is in practice more effective with the novel controllers not requiring CISs, we performed a statistical analysis on the sets found for the two setups of Section~\ref{ssec:simu_setpoint},~\ref{ssec:simu_disturbance} and~\ref{sec:experiments}.}

\review{We have tested different invariance conditions (listed in Fig.~\ref{fig:safe_sets}) on 100k states, sampled close to the set boundaries, as they are more likely to violate them. 
These tests boil down to solving one or multiple OCPs considering the conditions summarized in Section~\ref{ssec:sets_summary}. 
The percentage of states for which the solver fails to find a solution for a given condition is a measure of how far the set is to belong to the associated set class. }

\review{The results of this analysis are reported in Table~\ref{tab:set_4dof_res} and~\ref{tab:set_5dof_res}.
The horizon used to test classic Control Invariance was $45$ timesteps, for Up-to-step Control Invariance (\ref{eq:up_to_n_step_ci}) we tested all the horizons in $[2,45]$ and we reported the results for the one ($K$) that caused fewer solver failures, for Within-N-step Control Invariance (\ref{eq:weak_n_step_ci}) we reported the total number of state for which none of the horizons in $[2, 45]$ can satisfy the set condition. For back-and-forth-N-step Control Invariance (\ref{eq:baf_n_step_ci}) we always used as forward horizon $N=45$, reporting the results for the value of $k$ in $[2, 45]$ satisfying the set condition for more states. Finally, for back-and-forth-within-N-step Control Invariance (\ref{eq:condition_rf_pcmpc2}), we reported the number of states for which a combination $r$ and $j$ that satisfy the set condition does not exist.}

\review{Unsurprisingly, the two sets we tested are not exactly control invariant, and the invariance condition is violated in almost 1\% of the sampled states.
However, as we move towards less strict multi-step invariance definitions, the number of sampled states violating the invariance condition decreases by a factor of $\approx$45 for the first set and $\approx$10 for the second one.
This explains why the newly presented controllers perform better in practice, even though the used safe set is computed as the approximation of a classic CIS. Moreover, these results also support the claim that numerical procedures are more likely to produce a set that belongs to one of the presented classes.}

\begin{table}[tb]
\centering
\caption{\review{Statistical test results for the 4-DOF set used in Section~\ref{ssec:simu_setpoint}.}}
\rowcolors{2}{gray!10}{white}
\begin{tabular}{lc}
\toprule
\textbf{Set} & \textbf{Failures (\%)} \\
\midrule
CIS & 0.91 \\
Up-to-N-step CIS, [K = 3]  & 0.69 \\
Within-N-step CIS & 0.62 \\
Back-and-forth-N-step CIS, [r = 7] & 0.031 \\
Back-and-forth-within-N-step CIS & 0.02 \\
\bottomrule
\end{tabular}
\label{tab:set_4dof_res}
\end{table}

\begin{table}[tb]
\centering
\caption{\review{Statistical test results for the 5-DOF set used in Section~\ref{ssec:simu_disturbance}.}}
\rowcolors{2}{gray!10}{white}
\begin{tabular}{lc}
\toprule
 \textbf{Set} & \textbf{Failures (\%)} \\
\midrule
CIS & 0.72 \\
Up-to-N-step CIS, [K = 24] & 0.56 \\
Within-N-step CIS & 0.52 \\
Back-and-forth-N-step CIS, [r = 4] & 0.079 \\
Back-and-forth-within-N-step CIS & 0.071 \\
\bottomrule
\end{tabular}
\label{tab:set_5dof_res}
\end{table}
\section{Experiments}
\label{sec:experiments}

\begin{figure*}[tb]
    \centering
    \includegraphics[trim = 300 80 300 80, clip, width=0.19\textwidth]{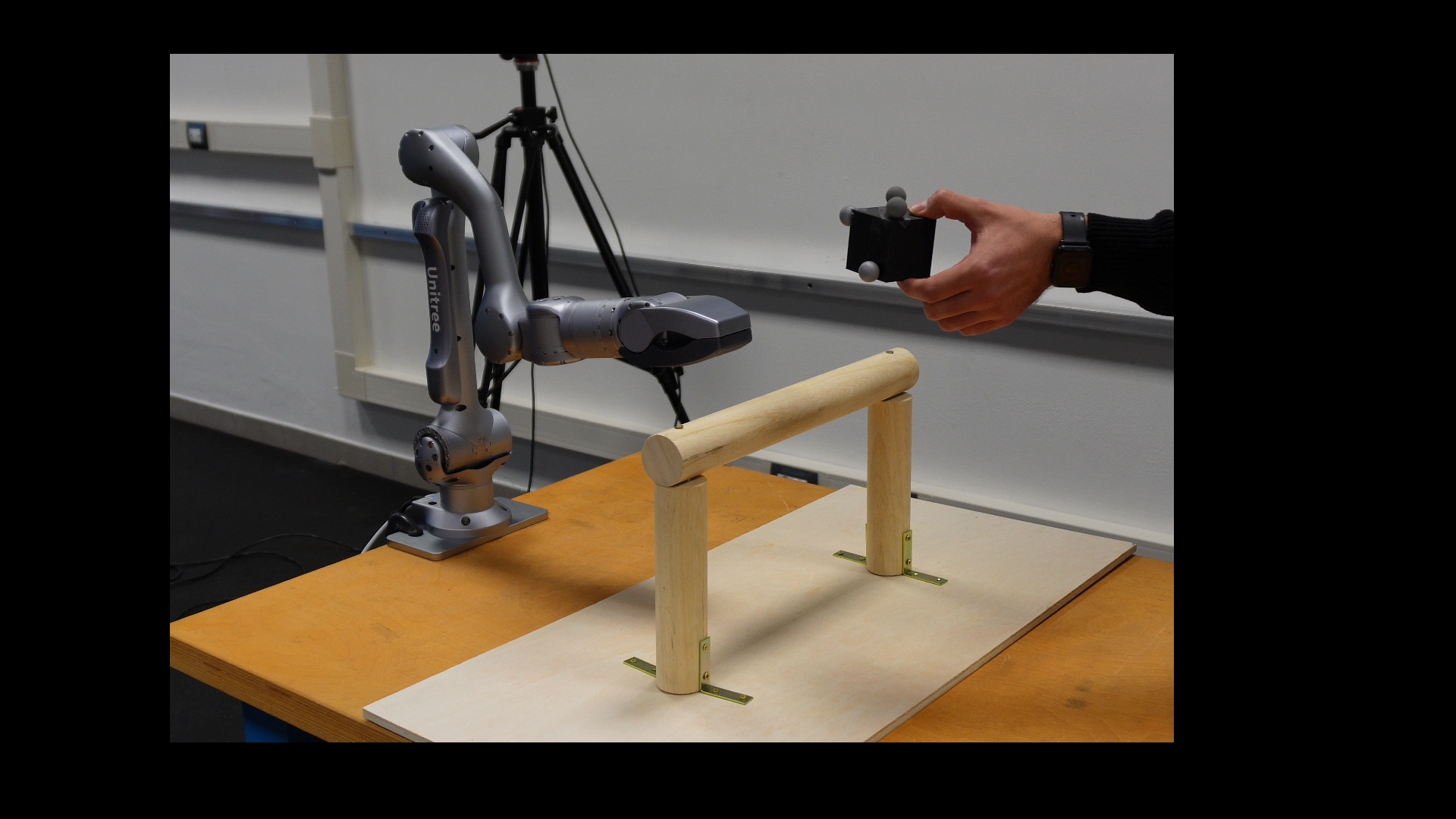} 
    \includegraphics[trim = 300 80 300 80, clip, width=0.19\textwidth]{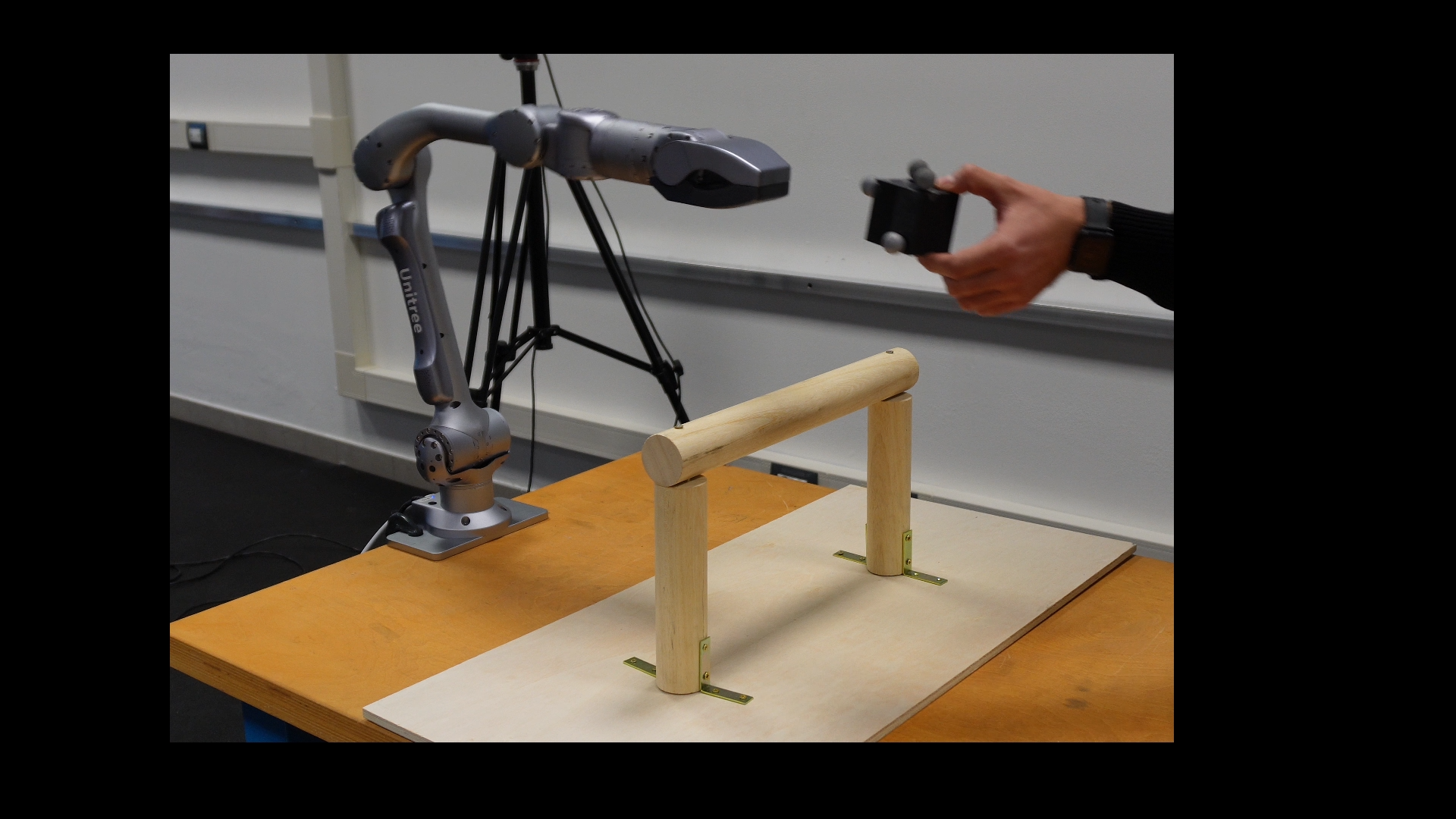} 
    \includegraphics[trim = 300 80 300 80, clip, width=0.19\textwidth]{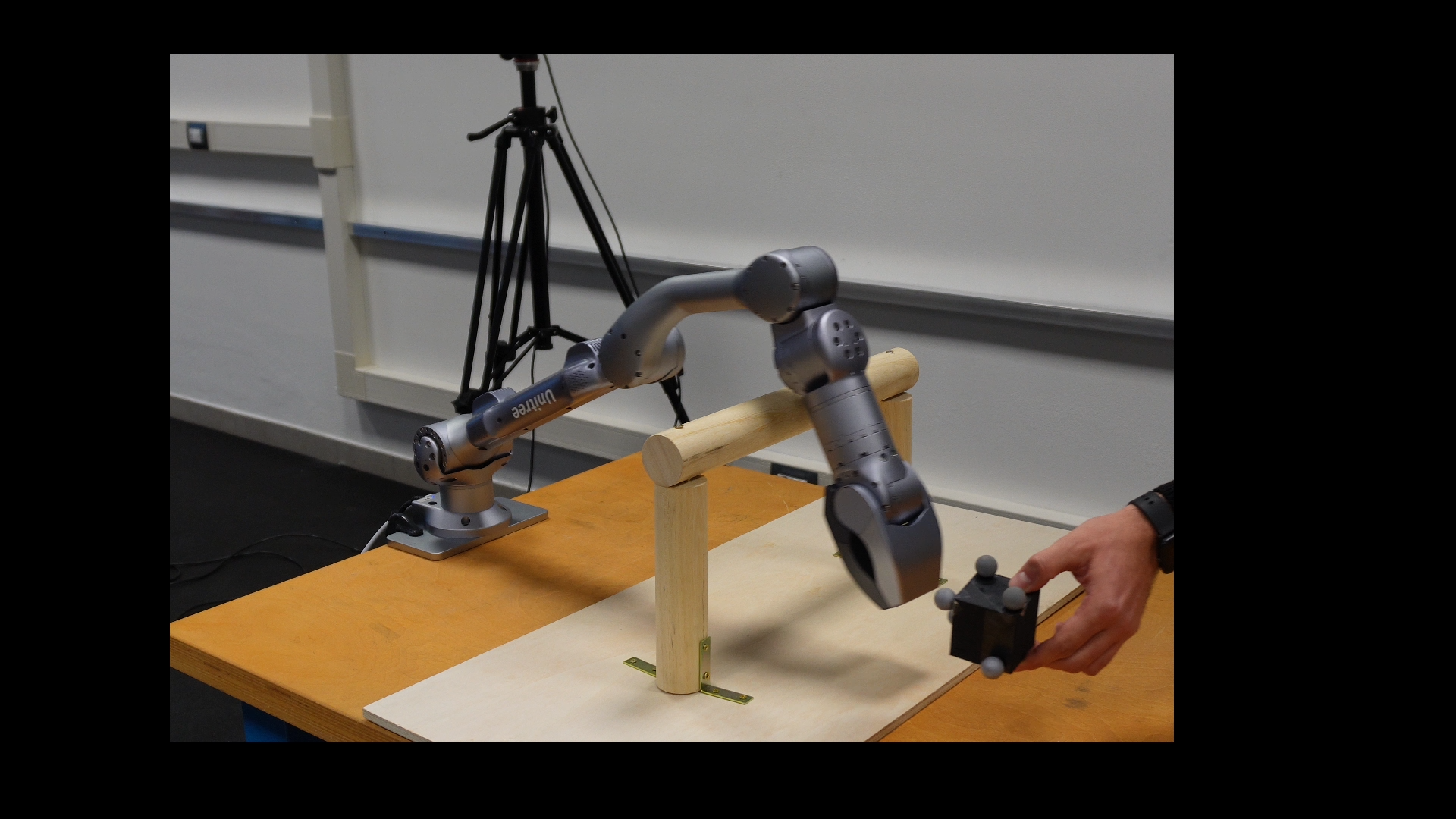} 
    \includegraphics[trim = 300 80 300 80, clip, width=0.19\textwidth]{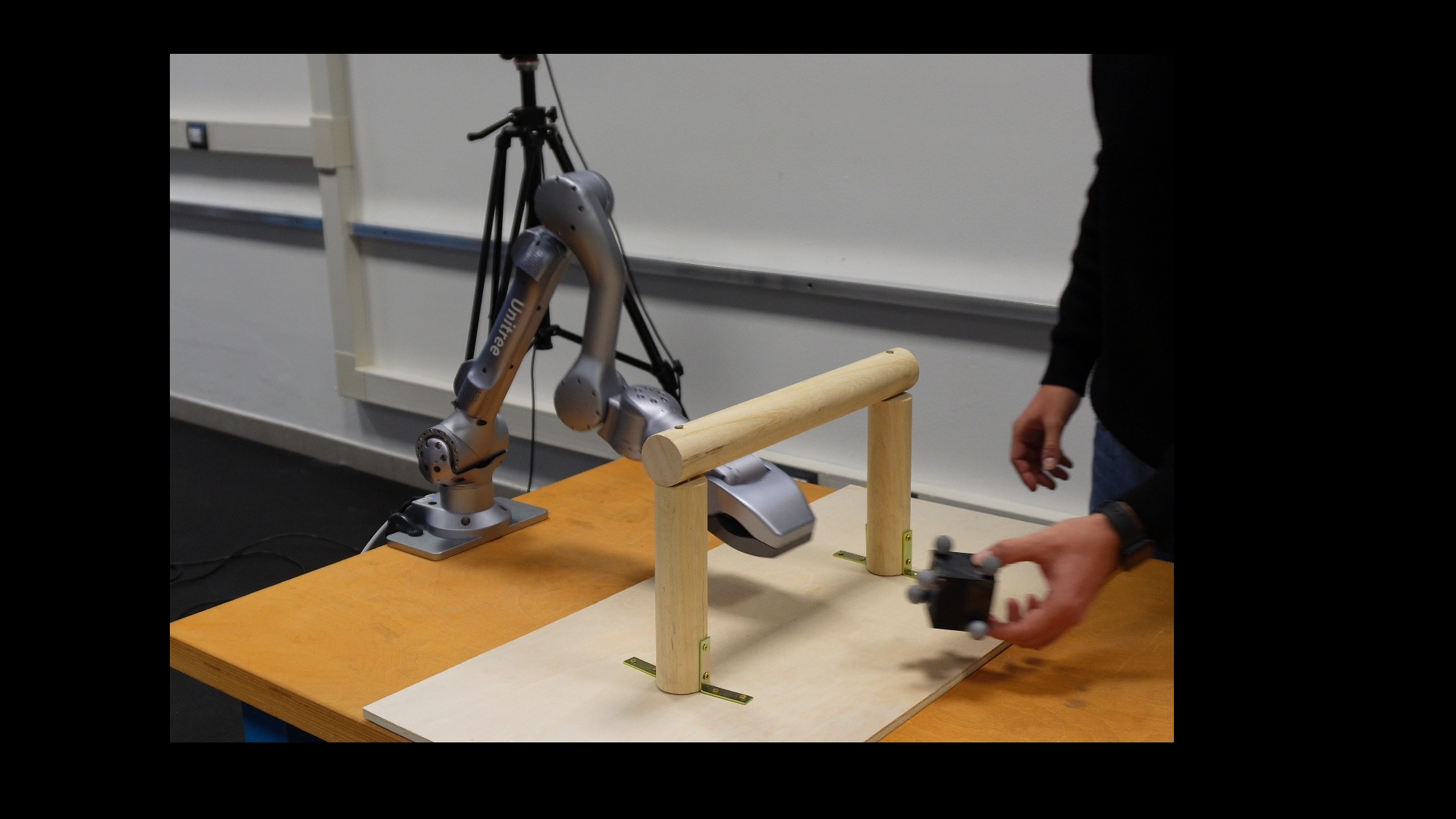} 
    \includegraphics[trim = 300 80 300 80, clip, width=0.19\textwidth]{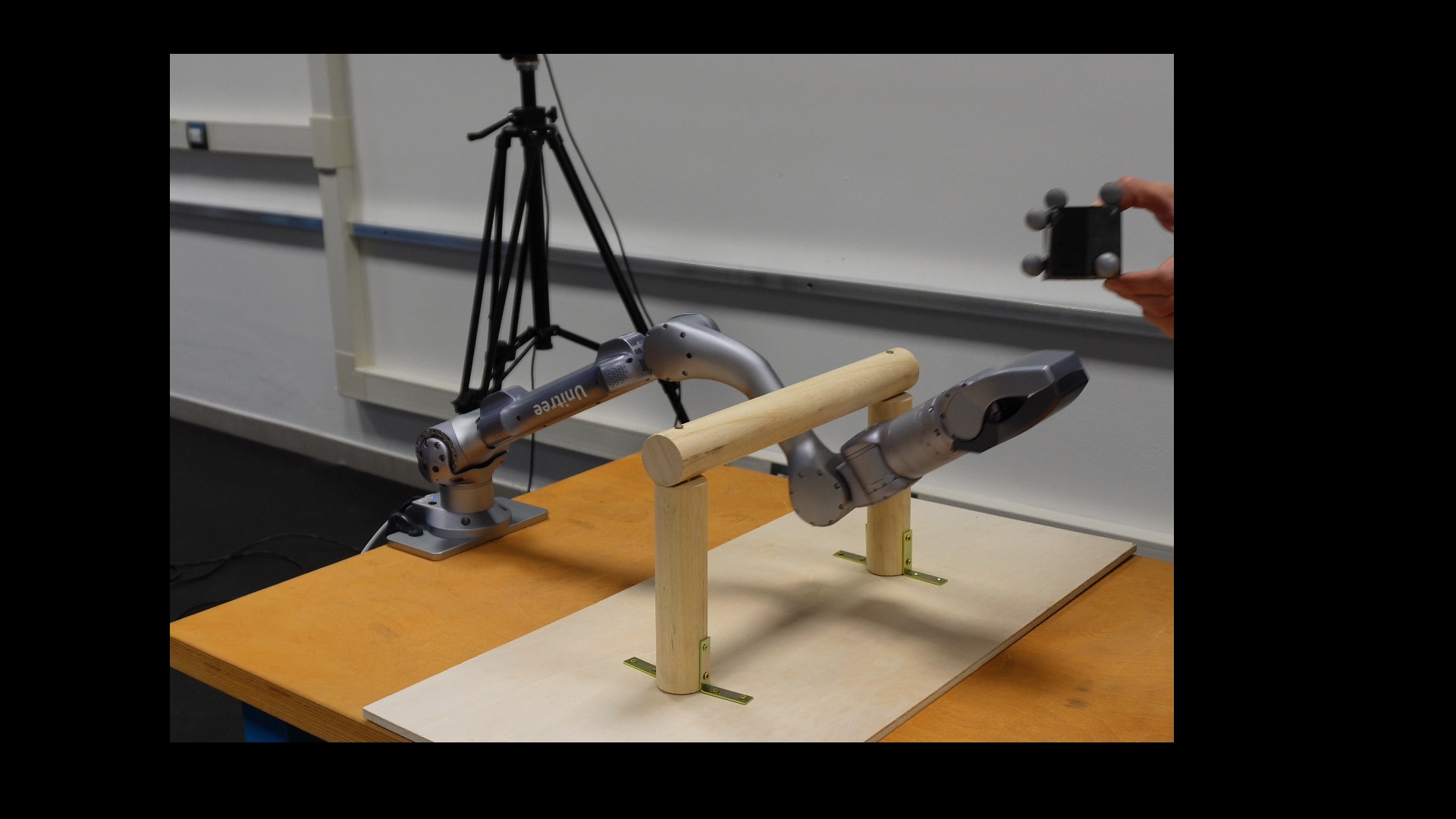} 
    \caption{Setup and example of robot motion with the RC-MPC controller (videos can be found in the supplementary material). \review{The manipulator tracks the moving target (the cube with markers in the human's hand) without colliding with the environment and without violating the joint limits. The frames, ordered in time from left to right, have been taken from a single experiment.}}
    \label{fig:z1_photos}
\end{figure*}

We validate one of the analyzed MPC approaches, namely the RC-MPC, on real hardware experiments, which are shown in the companion video. We perform the MPC on the full 6-DOF Z1 manipulator.
We were able to use a time step $\mathrm{d}t = \SI{10}{\milli\second}$ consistent with the simulations of Section~\ref{ssec:simu_disturbance}, despite the increased state dimension.
The setup and some frames of robot motion are reported in Fig.~\ref{fig:z1_photos}. The robot's EE must track the black box in the hand of the human, keeping a safety distance $p_{\mathrm{safe},x} = \SI{15}{\centi\meter}$ along the $x$-axis. The task has been defined with the following cost function:
\begin{equation}
    \begin{aligned}
        \ell(x_t, u_t) =& \norm{p_e(x_t)-p_e^\mathrm{ref}(t)}_{W_{t}}^2 +  \\
        & \norm{\mathrm{tr}(I_3 - R_\mathrm{ref}^\top \,R_e(x_t))}_{W_{r}}^2 + \norm{u}_{W_u}^2, 
    \end{aligned}
    \label{eq:exp_cost}
\end{equation}   
with:
\begin{equation}
     \begin{aligned}
        p_e^\mathrm{ref}(t) &= p_\mathrm{mocap}(t) - p_\mathrm{safe}, \quad R_\mathrm{ref} = I_3    \\
        W_t &= 100 \, I_3, \quad W_r = I_3, \quad W_u = 5 \cdot 10^{-3} \, I_6.
     \end{aligned}
\end{equation}
The position of the black box $p_\mathrm{mocap}(t)$ is provided by a motion capture system and it is updated with a frequency of $\SI{10}{\hertz}$. While tracking the target, the robot's EE must also maintain a reference orientation $R_\mathrm{ref}$, which is aligned with the global reference frame on the robot's base. 
The environment is the same as that shown in Fig.~\ref{fig:env_setup_disturbace}; therefore, we reused the NN trained on the 5-DOF system, adapting the safety constraint to the full DOF manipulator using~\eqref{eq:nn_constraint_mis} with $m = 5$. Due to the high joint friction of the Z1, we conducted the experiments using an open-loop MPC approach.
In this setup, the MPC re-computes the state and control trajectories at each time step based solely on its previous prediction, rather than incorporating real-time state feedback, following an approach similar to that proposed in~\cite{Neunert2017}. Specifically, we avoided closed-loop MPC execution, as the significant friction would interfere with accurate feedback control. Instead, at each time step, we sent the MPC state $x_t^*$ to the hardware using the Unitree\textregistered{} Z1 Python API. The robot then tracked this reference using its internal Proportional Derivative controller.

\subsection{Computation time}
\begin{figure}[tb]
    \centering
    \includegraphics[width=\columnwidth]{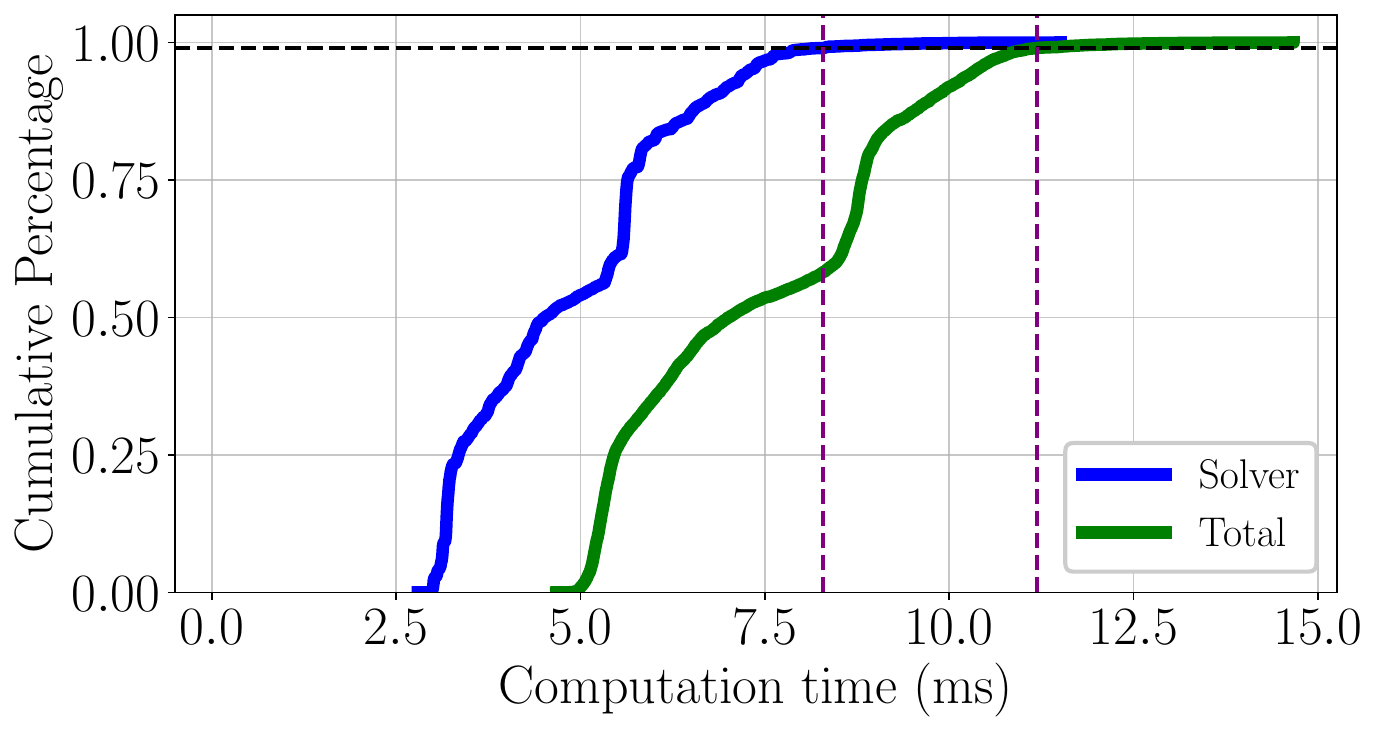}
    \caption{Cumulative percentage of the computation times of the RC-MPC during a real experiment. These plots show the likelihood of the computation time being less than a certain threshold.}
    \label{fig:exp_timings}
\end{figure}

Figure~\ref{fig:exp_timings} shows the computation time recorded during a single RC-MPC experiment. The blue curve refers to the time required by \textsc{Acados} for solving a single MPC iteration using the RTI scheme. The green curve represents the total time taken by the entire Python cycle, which includes the MPC computation, the reference target position update, and the verification checks on the next predicted states, such as bounds satisfaction and obstacle avoidance. A horizontal dashed line at the $0.99$ cumulative probability level marks the 99th percentile. The intersections of this line with the two curves define the upper bounds of the time intervals within which $99\%$ of the solver and total cycle durations fall, respectively of $\SI{8.29}{\milli\second}$ and $\SI{11.2}{\milli\second}$. The 99th percentile of the solver time is lower than the time step $\mathrm{d}t$, thus the MPC can be run at $\SI{100}{\hertz}$. Unfortunately, the requirement is not satisfied if we consider the total time required by the Python cycle, with $92\%$ of the iterations having a computation time below $\SI{10}{\milli\second}$. Still, we believe that an optimized implementation (e.g., using compiled programming languages such as C++) will largely decrease the gap between solver and total cycle durations.

\subsection{Constraint satisfaction}
\begin{figure}[tb]
    \centering
    \includegraphics[width=\columnwidth]{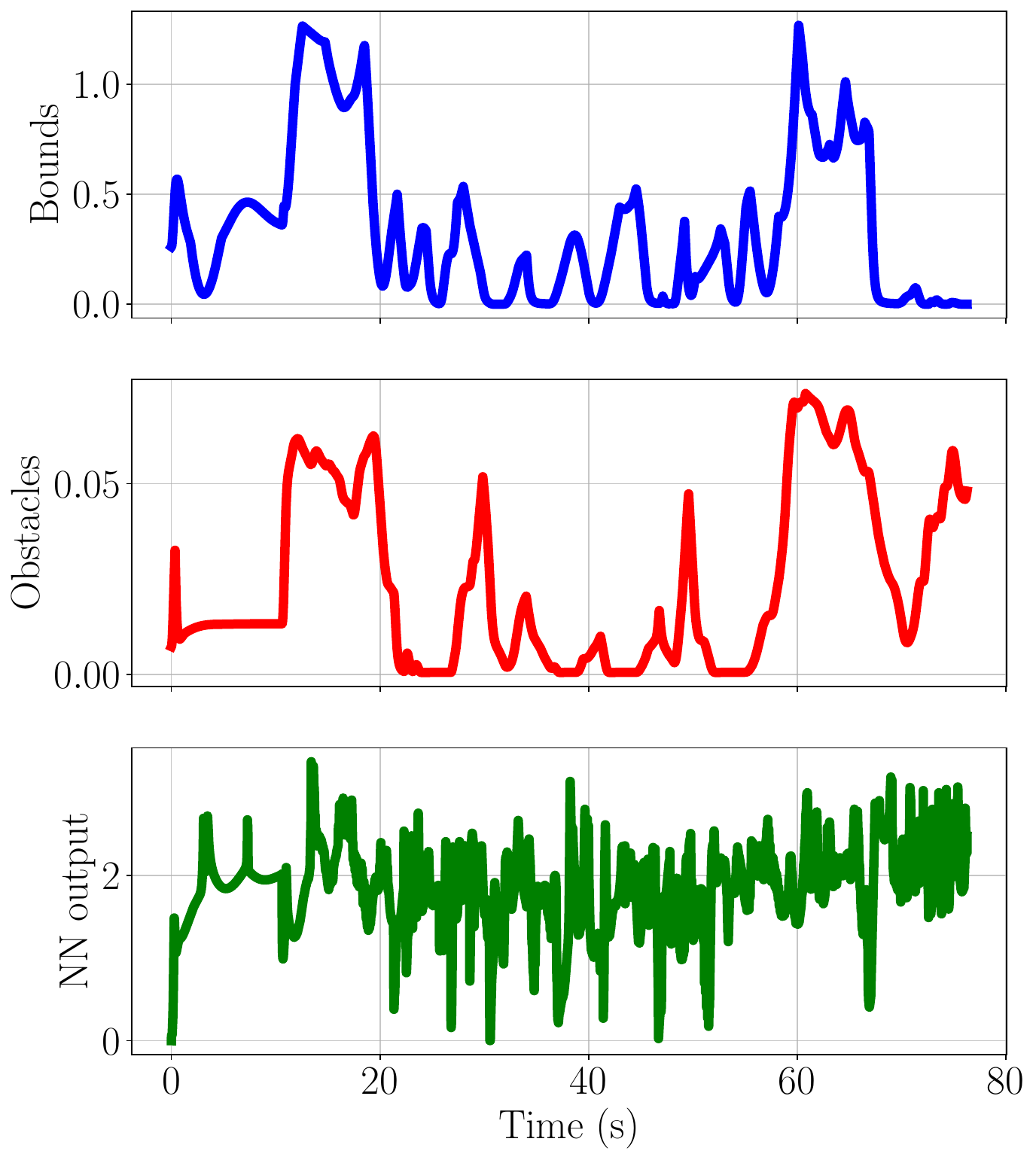}
    \caption{Constraint violations during the real experiment with RC-MPC.}
    \label{fig:exp_violations}
\end{figure}

As discussed in Section~\ref{ssec:rti_scheme}, constraint satisfaction must be verified after each MPC iteration. Figure~\ref{fig:exp_violations} illustrates the evolution of constraint violations related to state bounds, obstacle avoidance, and safe set membership according to the NN output. While the first two quantities depend on the current state at each sampling instant, the NN output is evaluated at the receding node $r$. Given the lower and upper state bounds $x_\mathrm{min}$ and $x_\mathrm{max}$, the bounds violation is defined as:
\begin{equation}
    \min\{x_\mathrm{max} -x, x - x_\mathrm{min}\}.
\end{equation}
Obstacle violation is quantified by computing the minimum distance between the robot and surrounding obstacles at each instant.

During the experiments, the robot manipulator is intentionally guided towards the obstacles through a controlled movement of the black box target, as shown in Fig.~\ref{fig:z1_photos}, to evaluate the robustness of the RC-MPC. All violation measures remain positive, indicating that the constraints are always respected.

\section{Conclusions}
\label{sec:conclusions}
In this paper, we tackled the challenge of ensuring safety in control strategies for robotics systems. We extended classical theory of control invariance by introducing novel classes of safe sets that generalize the traditional notion of CIS. These definitions allow for the design of more flexible MPC schemes, especially when dealing with approximate CIS in nonlinear systems.

Using some classes of safe sets, we have shown that some recently developed MPC schemes can guarantee RF. By moving beyond the classical terminal constraint paradigm and moving the safe set constraint across the prediction horizon, our schemes offer an adaptable framework for incorporating safety without highly diminishing the controller's performance.
To handle the inherent approximation of the safe sets, we introduced a \emph{safe abort procedure} that is triggered when a potential violation of constraints is recognized.

We validated the effectiveness of our methods through extensive simulations on robotic manipulators and demonstrated real-world applicability with hardware experiments on a 6-DOF robot arm. The results show that our approach reduces constraint violations compared to state-of-the-art methods, while maintaining acceptable tracking performance and computational efficiency.

The assumptions on the safe set \review{belonging to one of the multi-step classes} are still hard to verify in practice, as shown by the constraint violations in our tests. Inspired by recent work~\cite{Robey2020,Akella2023}, we want to investigate the use of robust optimization or probabilistic techniques to certify those sets. 
Another promising future work is the integration of our safety framework as a filter for data-driven policies, such as those trained with Reinforcement Learning.
Additionally, applying our safe MPC schemes to higher-dimensional and underactuated systems, such as legged robots, poses a relevant challenge. 
Using hardware accelerators for the computation of safe sets could help scaling to higher dimensionality. 
However, the challenges introduced by the non-smooth contact dynamics seem harder to overcome. 
Investigating the integration of our novel idea of multi-step safe sets with the framework of Model Predictive Shielding could be a possible solution, as it does not require differentiable dynamics. 
\review{Future work will investigate more flexible multi-step CIS representations, either by augmenting the system state with obstacle states, despite the resulting increase in computational cost, or by leveraging learning-based HJR methods that incorporate lidar measurements directly into the safe-set approximation~\cite{Lin2026}.}


\begin{acks}
Funded by the European Union. Views and opinions expressed are however those of the author(s) only and do not necessarily reflect those of the European Union or the European Health and Digital Executive Agency (HADEA). Neither the European Union nor HADEA can be held responsible for them. EU - HE Inverse - Grant Agreement 101136067.
\end{acks}

\bibliographystyle{SageH}
\bibliography{references.bib}

\appendix

\section{\review{Relationship between Multi-Step Control Invariance and Backward Reachability}}
\label{app:back_reach}
\review{We clarify here the relation between the multi-step control invariant sets introduced in this paper and classical backward reachable sets/tubes. Although these notions are related, they are not equivalent.}

\review{Given a target set $T \subseteq X$, an $N$-step backward reachable tube (BRT) characterizes the set of states from which the system can be driven to $T$ in at most $N$ steps. Similarly, an $N$-step backward reachable set (BRS) characterizes the set of states from which the system can be driven to $T$ in exactly $N$ steps. These constructions therefore require the prior specification of an external target set $T$.}

\review{By contrast, the multi-step control invariant sets considered in this paper are self-referential. In particular, a set $S$ is within-$N$-step control invariant if, from every state in $S$, there exists an admissible control sequence that drives the system back to $S$ within at most $N$ steps. Thus, the set to be reached is the same set whose invariance property is being assessed. Equivalently, using the notation of this paper, $S$ is within-$N$-step control invariant if
\[
S \subseteq \bigcup_{j=1}^{N} \mathcal{B}_j(S),
\]
where $\mathcal{B}_j(S)$ denotes the $j$-step backward reachable set of $S$.}

\review{This expression highlights the connection with reachability: backward reachable sets can be used as a tool to verify whether a given set $S$ satisfies the within-$N$-step control invariance property. However, the property itself is not simply an $N$-step BRT or BRS of a prescribed target set. In a standard BRT/BRS computation, the target set is fixed a priori and the reachable set is computed relative to that target. In the present setting, the candidate set $S$ must instead be contained in its own finite-step backward reachable tube. This self-consistency requirement is what defines the multi-step control invariance property.}

\review{Therefore, while reachability operators naturally appear in the characterization of the proposed set classes, the proposed multi-step control invariant sets are distinct from classical backward reachable sets and tubes. Classical BRT/BRS constructions may support the verification of the proposed properties for a given set, but they do not by themselves define or compute the self-referential multi-step control invariant sets introduced in this work.}


 





\end{document}